\documentclass[letterpaper,twocolumn,10pt]{article}
\usepackage{usenix-2020-09}

\usepackage{subfig}
\usepackage{xcolor}
\usepackage{listings}
\usepackage{booktabs}
\usepackage{graphicx}
\usepackage{amsmath}
\usepackage{makecell}
\usepackage{pifont}
\usepackage{amssymb}
\usepackage{multirow}
\usepackage{enumitem}
\usepackage{tikz}
\usepackage{xparse}
\usepackage{xspace}
\usepackage{etoolbox}
\usepackage{tcolorbox}
\usepackage{textcomp} %
\usepackage{comment}
\usepackage{inconsolata} %
\usepackage{underscore} %
\usepackage{tabularx}
  
\usepackage[english]{babel}

\begin{document}

\title{\Large \bf BLens: Contrastive Captioning of Binary Functions using Ensemble Embedding}

\author{
{\rm %
Tristan Benoit$^{*\dagger\mathparagraph}$, 
Yunru Wang$^{*\dagger\ddagger}$, 
Moritz Dannehl$^{\dagger\ddagger}$, 
Johannes Kinder$^{\dagger\ddagger}$}\\[6pt]
$^\dagger$ Ludwig-Maximilians-Universität München, Germany\\
$^\ddagger$ Munich Center for Machine Learning, Germany\\
$^\mathparagraph$ Bundeswehr University Munich, Germany
} 

\definecolor{mGreen}{rgb}{0,0.6,0}
\definecolor{mGray}{rgb}{0.5,0.5,0.5}
\definecolor{mPurple}{rgb}{0.5,0,0.8}
\definecolor{mMagenta}{rgb}{0.7,0,0.6}
\definecolor{backgroundColour}{rgb}{0.96,0.96,0.96}

\lstdefinestyle{CStyle}{
    basicstyle=\ttfamily\footnotesize,
    backgroundcolor=\color{backgroundColour},   
    commentstyle=\color{mGreen},
    keywordstyle=\color{mMagenta},
    numberstyle=\tiny\color{mGray},
    stringstyle=\color{mGray},
    breakatwhitespace=false,         
    breaklines=true,                 
    captionpos=b,                    
    keepspaces=true,                 
    numbers=left,                    
    numbersep=5pt,                  
    showspaces=false,                
    showstringspaces=false,
    showtabs=false,                  
    tabsize=2,
    language=C,
}

\newcommand{\ProjectName}{BLens\xspace}

\newcommand{\TP}{\mathit{TP}}
\newcommand{\FP}{\mathit{FP}}
\newcommand{\FN}{\mathit{FN}}

\def\Snospace~{\S{}}
\renewcommand*{\sectionautorefname}{\Snospace}
\renewcommand*{\subsectionautorefname}{\Snospace}
\renewcommand*{\subsubsectionautorefname}{\Snospace}

\let\OLDthebibliography\thebibliography
\renewcommand\thebibliography[1]{
  \OLDthebibliography{#1}
  \setlength{\parskip}{0pt}
  \setlength{\itemsep}{2pt plus 0.3ex}
}

\newcommand{\myparagraph}[1]{{\textbf{{#1}.}}}

\NewDocumentCommand\casebullet{m}{%
    \tikz[baseline=(char.base)]{
        \node[shape=circle, draw=black, thick, fill=black, inner sep=1pt] (char) {\textcolor{white}{\small \sf #1}};
    }%
}

\newtcolorbox{mybox}[1]{colframe=white!20!black,title=#1,boxsep=1pt,left=5pt,right=5pt,top=5pt,bottom=5pt}

\maketitle

\begingroup
  \renewcommand\thefootnote{$*$}
  \footnotetext{Both authors contributed equally to this work.}
\endgroup

\begin{abstract}
Function names can greatly aid human reverse engineers, which has spurred the development of machine learning-based approaches to predicting function names in stripped binaries. Much current work in this area now uses transformers, applying a metaphor of machine translation from code to function names. Still, function naming models face challenges in generalizing to projects unrelated to the training set. 
In this paper, we take a completely new approach by transferring advances in automated image captioning to the domain of binary reverse engineering, such that different parts of a binary function can be associated with parts of its name.
We propose BLens, which combines multiple binary function embeddings into a new ensemble representation, aligns it with the name representation latent space via a contrastive learning approach, and generates function names with a transformer architecture tailored for function names.
Our experiments demonstrate that BLens significantly outperforms the state of the art. In the usual setting of splitting per binary, we achieve an $F_1$ score of 0.79 compared to 0.70. In the cross-project setting, which emphasizes generalizability, we achieve an $F_1$ score of 0.46 compared to 0.29. Finally, in an experimental setting reducing shared components across projects, we achieve an $F_1$ score of $0.32$ compared to $0.19$.
\end{abstract}

\section{Introduction}

Reverse engineers analyze programs available only in binary form in security audits, for vulnerability discovery, or during forensic analysis of malware~\cite{Cifuentes99,bughunting02,sikorski-practicalmalware}. Disassemblers such as IDA Pro or Ghidra provide assistance by discovering functions within the binary and resolving data references and control flow. Programs being analyzed have usually been stripped of most human-readable information, such as meaningful function names. Because function names help to effectively navigate the code and identify points of interest, human reverse engineers are known to manually label functions in the disassembly with names that capture their semantics~\cite{VotipkaRMFM20,Mantovani22}.

Machine learning promises to help automate tasks that require a human level of understanding, especially with imprecise concepts such as developer-assigned names. 
Learned embeddings of binary code aim to capture the semantics of assembly code in a compact vector representation~\cite{asm2vec,alphadiff,safe,gemini,palmtree,WangISSTA24,wang2022jtrans}. Binary code embeddings can be used directly for binary code similarity detection or as part of more complex architectures and tasks.

Pioneering work on naming binary functions aims to learn and predict complete function names~\cite{he2018debin,acsac20-punstrip,han21issta}, an approach which is by design limited to a coarse granularity of semantics and mostly recovers function names frequently seen in the training data~\cite{oakland23-xfl}. 
Splitting function names into sets of labels or tokens addresses these limitations~\cite{nero,oakland23-xfl}. However, a multi-label setting ignores the ordering of words inside a function name, and ordering during post-processing misses important nuances in function names.

Recent transformer-based approaches use an encoder-decoder architecture~\cite{transformer} to effectively treat function naming as a translation problem from binary code to function names, which are seen as sentences of tokens~\cite{jin22ccs, kim23asiaccs,nero,XiongASE23}. 
A key design choice in this line of work lies in the binary code representation available to the encoder. For instance, AsmDescriptor~\cite{kim23asiaccs} works on assembly code, while HexT5~\cite{XiongASE23} leverages a decompiler to work directly on pseudocode. %

In this paper, we argue that translation is \textit{not} the right metaphor in function name prediction but that the task is closer to generating meaningful captions for images~\cite{ke2019reflective}. In particular, our intuition is that \textit{binary code and function names should be treated as two modalities for the same concept}.

By integrating ideas from the multimodal machine learning field~\cite{baltruvsaitis2018multimodal}, we aim to obtain a solid relation between parts of a function and words. For instance, Contrastive Language-Image Pre-Training (CLIP)~\cite{clip} consists of breaking down images into smaller patches and then associating them with corresponding human-readable text during pre-training. 
Vision transformers~\cite{vit} process image patches in a way that captures spatial relationships between them.
Just as image captioning requires understanding different visual components (e.g., colors and shapes) and their visual relationships to generate a coherent description, function name prediction involves associating parts of binary code with function names in a way that captures interrelationships in the binary code.
These relationships reside at the level of control and data flows, as well as within operations, function calls, and resources, such as strings.
Still, there are differences between images and functions that one should consider.
First,  it is natural to obtain image patches by cutting images, but a function can be characterized by various structures, from sequences to graphs. Second, currently available function datasets comprise millions of functions at most, while image datasets comprise billions of images~\cite{zhang2021vinvl}.
Third, while results for function naming published so far are promising, they still do not generalize well across separate projects, as we explain below.
\myparagraph{Distribution shifts}   Most work in learning-based binary analysis adopts the \mbox{\bf cross-binary setting}, where functions in the training, validation, and test sets are drawn from distinct binaries. %
Xiong et al.~\cite{XiongASE23} argue that the {\bf cross-project setting}, where an entire project is assigned to a single set,  better reflects real-world use cases, as reverse engineers typically encounter binaries from unknown projects. This setting is difficult for learning-based methods as it demands better generalization capability from models. Binary code within the same project tends to have similar distributions due to shared components, coding practices, and compilation configurations, which may lead to overfitting. Consequently, accuracy drops significantly in the cross-project setting, as the training and test sets exhibit substantial distribution differences, referred to as {\bf distribution shifts}. While different projects may still share individual source code components or snippets, this is expected even in real-world scenarios. However, to evaluate solely generalization capabilities, we additionally introduce a strict evaluation setting (\mbox{\autoref{sec:strict_setting}}), in which we aggressively minimize the impact of shared components across projects.

\myparagraph{Challenges} We identify the following challenges to function name prediction on stripped binary code:
(C1)~Function names often encapsulate semantics sensitive to the order of words (e.g., \texttt{int\_to\_float}). It is thus important to capture word order through the binary code structure. 
(C2)~Both precision and recall are crucial for providing useful function names. A good precision guarantees that suggested names are relevant and not misleading, 
while a good recall ensures the model proposes names for most functions. Achieving a satisfactory amount of both is challenging.
(C3) Ensuring robust generalization is critical to the more challenging cross-project setting. Yet existing approaches are seldom evaluated in this realistic setting~\cite{oakland23-xfl,acsac20-punstrip,nero,han21issta,jin22ccs, he2018debin}.
\myparagraph{Our approach} We introduce \ProjectName (\textbf{B}inary \textbf{Lens}), which aligns the modality of binary code---represented as function patches---with human-readable text through the contrastive captioning (CoCa)~\cite{coca} multi-task. To obtain these patches, \ProjectName leverages state-of-the-art embeddings: \textsc{\textsc{Clap}}~\cite{WangISSTA24} for a cross-modal function embedding, \textsc{PalmTree}~\cite{palmtree} to represent sequences of basic blocks, and \textsc{Dexter}~\cite{oakland23-xfl} for a context-aware function embedding.
Because multi-task models are robust\mbox{~\cite{caruana1997multitask}} but suffer from conflicts among training objectives\mbox{~\cite{sener2018multi}}, we fine-tune BLens's decoder strictly for captioning.
In order to handle rather small datasets and short function names, the decoder learns through a Masked Language Modeling (MLM) task\mbox{~\cite{devlin-bert}} tailored to function names, and to maintain precision, it employs a confidence threshold. 
In particular, we make the following contributions:

\begin{itemize}
\item We present \ProjectName, a new approach to function name prediction inspired by multimodal machine learning~(\autoref{sec:mmml}). Our architecture integrates multiple existing binary code representations thanks to the  \textsc{Combo} (\textbf{CO}ntrastive \textbf{M}ulti-modal \textbf{B}inary embedding \textbf{O}ptimizer) pre-training phase (\autoref{sec:cc}) to achieve better generalization. %
\item We introduce \textsc{Lord} (\textbf{L}ikelihood \textbf{O}rdered \textbf{R}egressive \textbf{D}ecoder)~(\autoref{sec:lord}), a decoder and inference scheme that employs a hard MLM task for a more profound fine-tuning process along with a flexible autoregression.  \textsc{Lord} maintains consistently high precision even in the challenging cross-project setting.
\item  \ProjectName achieves substantial improvements over the state of the art, particularly on the grammatical structure. In an evaluation in the cross-binary setting~(\autoref{sec:crossbinarysetting}), we observe gains of 12\% $F_1$, 32.6\% RougeL, 79.1\% Bleu, and 11.1\% VarCLR scores. In the cross-project setting~(\autoref{sec:crossprojectsetting}), we observe gains of 42.2\% $F_1$, 71.7\% RougeL, 188\% Bleu, and 12.1\% VarCLR scores.
In the strict setting, which removes shared components\mbox{~(\autoref{sec:strict_setting})}, BLens provides a gain of $53.3$\% in terms of $F_1$ score.
An ablation study~(\autoref{sec:ablation}) validates our contributions by demonstrating a 55\% increase in $F_1$ score from the  \textsc{Combo} pre-training and a 56.2\% boost in precision from the \textsc{Lord} decoder.
\end{itemize}

Additionally, a case study of predictions in the cross-project setting~(\autoref{sec:discussion}) illustrates \ProjectName{}'s capacity to generate relevant function names that may differ from the ground truth, thereby addressing the lower $F_1$ score in this hard setting.

\ProjectName{}, along with all experimental data, is available as open source~\cite{blens-artifact}.

\begin{figure*}[h]
    \centering
    \includegraphics[width=\linewidth]{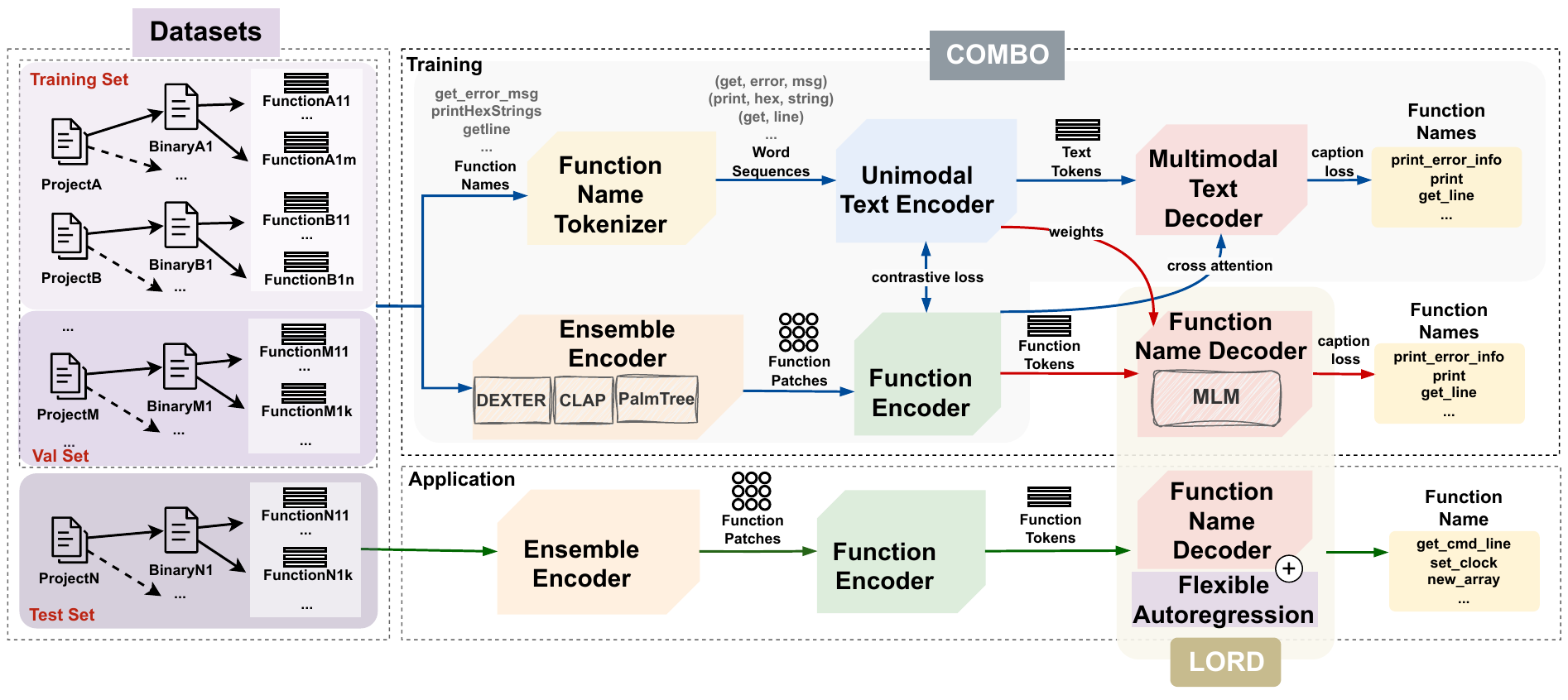}
    \caption{Overview of \ProjectName. Inspired by CoCa~\cite{coca} and GIT~\cite{git}, \ProjectName\ employs a two-stage process: pre-training (blue arrows) and fine-tuning (red arrows). In pre-training, we initialize function tokens using an ensemble encoder and pre-train the function encoder with a contrastive captioning task. During fine-tuning, the pre-trained encoder, combined with a function name decoder trained with a new MLM task, generates function names. In inference (green arrows), the model takes functions as input and outputs their corresponding names with the flexible autoregression.}
    \label{fig:arch}
\end{figure*}

\section{Multimodal Machine Learning\label{sec:mmml}}
This section introduces two approaches from multimodal machine learning for image captioning, CoCa~\cite{coca} and GIT~\cite{git}, which form the foundation of \ProjectName. As both are built around the transformer~\cite{transformer} as their core component, we begin by recalling its technical details.

\myparagraph{Transformer} The transformer~\cite{transformer} %
has recently found widespread application
in binary analysis~\cite{binbert, palmtree, jin22ccs, ordermatters, pei21arxiv, wang2022jtrans, pei2021stateformer, zhu2023ktrans, DBLP:conf/icml/PengZLKHL21, zhang2020similarity}, demonstrating its effectiveness in learning binary representations. It consists of an encoder and a decoder, each with several stacked blocks. The main components of encoder (and decoder) blocks are multi-head attention, residual connections~\cite{resnet}, and layer normalization~\cite{layernormalization}. Multiple attention heads perform self-attention independently, allowing the model to capture a wider range of dependencies by attending to different aspects of the input sequence simultaneously. Residual connections, implemented by adding the layer input directly to its output, help maintain gradient flow and prevent degradation in deep networks. Layer normalization, which operates by normalizing the summed inputs across each neuron before applying activation, ensures consistent activation values and stabilizes training.%

\myparagraph{CoCa} CoCa~\cite{coca} includes an image encoder, a unimodal text decoder, and a multimodal text decoder. The image encoder processes image patches, while the uni- and multimodal decoders form the two halves of a transformer decoder. The unimodal decoder encodes text using masked self-attention, and the multimodal decoder combines the outputs of the unimodal decoder and the image encoder with cross-attention to create image-text representations. The objective of CoCa combines contrastive and captioning losses. The contrastive loss aligns text and image tokens in the same latent space, while the captioning loss is a generative loss derived from predicting captions.

\myparagraph{GIT} Compared to CoCa, GIT~\cite{git} is a simpler generative model for image/video captioning, comprising only one image encoder and a text decoder. It concatenates image tokens with text tokens as input to the transformer module and predicts the associated description for the input image. Notably, GIT employs an existing encoder pre-trained with  a contrastive pre-training task~\cite{yuan2021florence} as the image encoder, similar to how we pre-train a function encoder based on the contrastive captioning (CoCa) task.

\section{Overview}

Now, we first define the problem and then introduce the training and application process of our model.

\myparagraph{Problem Definition} Function name prediction automatically recovers function names from stripped binary code. Formally, the process of recovering function names from functions is defined as
\[
\mathit{Name}_i = G(F_{i}^{p, b})
\]
where \( F_{i}^{p, b} \) represents the \( i \)-th function extracted from binary \( b \) in project \( p \), \( G \) denotes our model, \( \mathit{Name}_i \) is the ground truth name for \( F_{i}^{p, b} \). The input of the model is a function, encompassing related code, data flow, control flow, and other features. The output of the model is a sequence of words \( \mathit{\hat{Name}}_i \in \mathit{Vocab}^{n} \)  where \( \mathit{Vocab} \) is the set of words from the vocabulary and \( n \) is the sequence length. The effectiveness of the model is judged based on how closely its output aligns with the ground truth.

\myparagraph{Terminology} We draw inspiration from standard terms used in image captioning tasks to denote intermediate outputs. Similar to how image patches are inputs to the image encoder, we refer to the inputs of the function encoder as function patches. Moreover, the outputs of encoders are termed tokens. Finally, each token is a vector of fixed dimensionality $d$. 

\myparagraph{Training stages} Training consists of a pre-training phase followed by a fine-tuning phase. During pre-training, \textsc{Combo} learns robust function representations (with blue arrows in \autoref{fig:arch}) with a contrastive captioning task. During fine-tuning, the new decoder \textsc{Lord} is added on top of the pre-trained function encoder to generate function names (with red arrows in \autoref{fig:arch}). Throughout the entire training process, all weights are updated.

\myparagraph{\textsc{Combo}} Firstly, given \( F_{i}^{p, b} \) as input, the ensemble encoder embeds processed function features with three state-of-the-art embeddings and then generates function patches.
Secondly, the tokenizer turns the ground truth name into the sequence of words $\mathit{Name}_i$ that the unimodal text encoder converts into text tokens.
Simultaneously, the function encoder transforms the function patches into function tokens. There, a contrastive task relates the outputs of both encoders. Thirdly, a multimodal text decoder integrated on top of both encoders generates multimodal text tokens. Lastly, a captioning task relates $Name_i$ to these tokens. Details are provided in \autoref{sec:cc}.

Previous contrastive pre-trainings on functions\mbox{~\cite{WangISSTA24,zhang2024contrabin}} do not combine contrastive and captioning losses, missing out on the advantages of a multi-task approach\mbox{~\cite{caruana1997multitask}}.
Additionally, they require robust summaries generated by LLMs as anchors, which are challenging to obtain\mbox{~\cite{goldman2024context}}. In contrast, we directly learn text tokens as anchors through the unimodal text encoder.

\myparagraph{\textsc{Lord}} Training on multiple tasks introduces conflicts\mbox{~\cite{sener2018multi}}. Therefore, we introduce a fine-tuning stage with a function name decoder, \textsc{Lord}, that is strictly focused on captioning through a novel Masked Language Modeling (MLM) task.
This task replaces the standard teacher forcing in transformers with predicting randomly selected words from the remaining context.  Additionally, during application, the task allows \textsc{Lord} to employ a flexible autoregression process, which picks probable words one step at a time. The flexible autoregression stops when the confidence scores of the proposed words fall below a fine-tuned threshold. The details are shown in \autoref{sec:lord}.

\myparagraph{Application} In \autoref{fig:arch}, the dashed box and green arrows at the bottom illustrate the actual application of the model.
Given \( F_{i}^{p, b} \) as input, the ensemble encoder begins by embedding processed function features with three state-of-the-art embeddings.
Then, it generates function patches from the initial embeddings.
Moreover, these patches are then transformed into function tokens by the function encoder.
Finally, following the flexible autoregression process, the function name decoder generates text tokens one at a time, which are then converted into a word sequence, resulting in \( \mathit{\hat{Name}}_{i} \).

By using the idea of image captioning, we capture the word order through the binary code structure, addressing challenge \textbf{C1}.
With \textsc{Combo}, we aim to address distribution shifts, mentioned in challenge \textbf{C3}, by associating function parts with similar semantics to corresponding words in names.
The \textsc{Lord} task and autoregressive process significantly boost precision (see \autoref{sec:lordeva}), which is key to challenge \textbf{C2}.

\section{\textsc{Combo}}\label{sec:cc}

We now present \textsc{Combo} in detail. With stripped binaries as input, the ensemble encoder outputs function patches using existing SotA approaches (\autoref{sec:efe}). By utilizing these existing models instead of building a custom binary representation model, we can leverage the diverse binary features extracted by different models, thereby enriching the representation and enhancing robustness. Next, we refine the function patches with a contrastive captioning task (\autoref{sec:dd}), which links function tokens to corresponding text tokens. The robust semantic understanding of \textsc{Combo} mitigates distribution shifts~\cite{distribution}.

\subsection{Ensemble Encoding}\label{sec:efe}
\begin{figure}
    \centering
    \includegraphics[width=\linewidth]{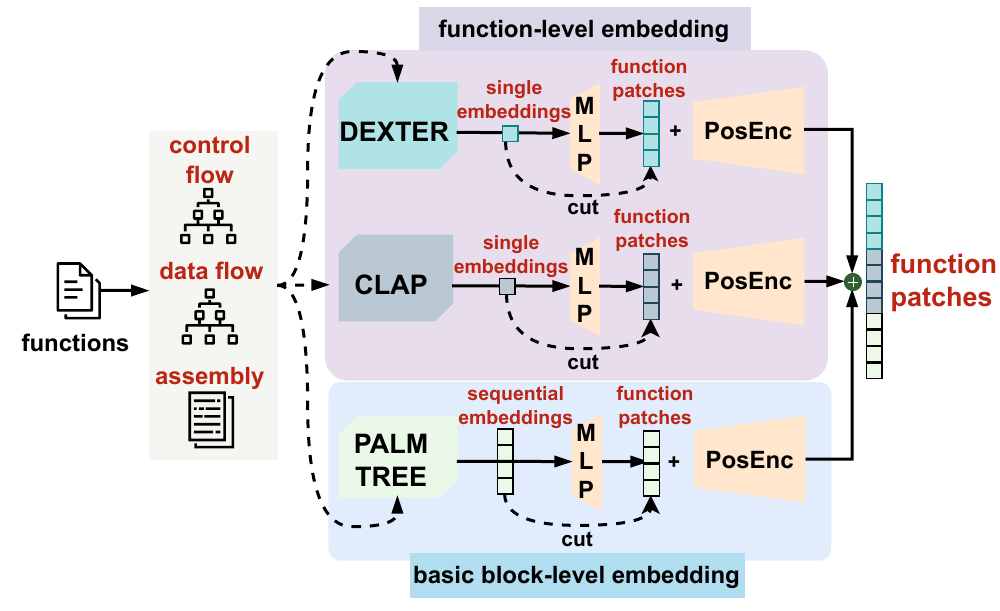}
    \caption{Overview of the ensemble encoder.}
    \label{fig:fe}
\end{figure}

To better represent functions, \ProjectName leverages knowledge from existing binary representation models instead of training new ones from scratch. Different models use distinct binary pre-processing techniques and extract unique features from functions. Thus, selecting suitable combinations is crucial to avoid redundancy and ensure complementary features.

We categorized state-of-the-art binary representation models by the granularity of their embeddings. Some approaches generate embeddings at the basic block level, while others produce a single embedding for an entire function. Based on the differences in embedding granularity, extracted binary features, accessibility, and performance, we selected \textsc{PalmTree}~\cite{palmtree}, \textsc{\textsc{Clap}}~\cite{WangISSTA24}, and \textsc{Dexter}~\cite{oakland23-xfl}. Although we focused on these three models, our module can be applied to various embedding approaches.

For basic block embeddings, we select \textsc{PalmTree}~\cite{palmtree}, a pre-trained BERT-based~\cite{devlin-bert} assembly language model. It generates instruction embeddings through self-supervised learning incorporating both data-flow and control-flow information, allowing for straightforward extraction of basic block embeddings. For function-level embeddings, we select \textsc{\textsc{Clap}} and \textsc{Dexter}. \textsc{\textsc{Clap}}~\cite{WangISSTA24} is the best-performing approach we have identified in our experiments, effectively aligning function code with descriptive text. \textsc{Dexter}~\cite{oakland23-xfl} employs static analysis to extract both quantitative (e.g., number of transitively reachable functions) and categorical features (e.g., MinHash hashes of assembly opcodes) from functions and uses MLP layers to integrate these features and produce function embeddings. Compared to most methods that use learning-based tasks to capture binary features, \textsc{Dexter} employs complex static analysis. It is also one of the best-performing binary embeddings for function name prediction.
The three models have distinct focuses: \textsc{PalmTree} emphasizes control flow and data flow dependencies among assembly instructions, \textsc{\textsc{Clap}} incorporates source code information, and \textsc{Dexter} focuses on statistical features.

Our attention mechanism requires many input patches to discern relationships effectively. While images, videos, and speeches can easily be cut into a sequence\mbox{~\cite{vit,chen2023xllm,lei2021lessismore}}, this is not the case for function-level embeddings.
A projection\mbox{~\cite{venugopalan2021multimodal,francis2019fusion, li2019graph2seq}} either fails to produce enough patches or requires so many parameters that it dilutes the effectiveness of the attention mechanism.
To effectively combine these embeddings, we use neural networks structured as in \autoref{fig:fe}, where PosEnc stands for the positional encoding. For function-level embeddings, we apply the concept of image patches, cutting embeddings into smaller embeddings and projecting them to function patches by an MLP layer. For basic block-level embeddings, an MLP layer projects them to patches. Learnable positional encodings are then added to each patch. Notably, after a function-level embedding has been turned into patches, the positional encoding of each patch indicates which part of the original embedding it contains. Finally, all patches are concatenated for joint training on the contrastive captioning task during pre-training.

\subsection{Contrastive Captioning}\label{sec:dd}
\begin{figure}
    \centering
    \includegraphics[width=\linewidth]{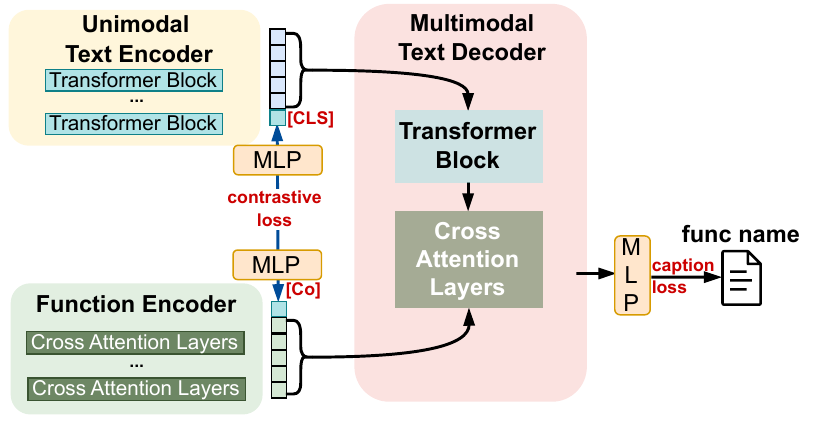}
    \caption{Overview of the contrastive captioning task.} %
    \label{fig:CoCa}
\end{figure}

Function name captioning can also serve as a pre-training task, as in CoCa~\cite{coca}. Contrastive captioning is the backbone of our robust modality translation from binary code to text.
In this section, we outline the detailed process of the contrastive captioning task.

Inspired by CoCa, we use a unimodal text encoder to encode words into text tokens and a function encoder to generate function tokens. By applying contrastive loss, we align the text tokens with function tokens. Additionally, we add a multimodal text decoder on top of the dual encoders to perform the captioning task, further enhancing the ability of the model to generate accurate and relevant function names.

The structures of the encoders and the decoder are shown in \autoref{fig:CoCa}.
In the dual-encoder setup, given a function and its name, the name is first tokenized into words and then embedded into $n+1$ initial text tokens, where $n$ is the maximum number of words in a name (padded if fewer), and the +1 accounts for the addition of a \texttt{[CLS]} token at the end of each token sequence. These initial text tokens serve as the input for the unimodal text encoder, which produces unimodal text tokens through transformer blocks.

For the function encoder, the function is first transformed into $k_1$ function patches of dimension $d$ using the ensemble encoder. Remember that $d$ is also the dimension of each token. Function patches are then passed as input to the function encoder, which outputs  $k_2$ function tokens through cross-attention with $k_2$ learnable queries. Among these function tokens, one is reserved for the contrastive task, referred to as the contrastive token, while others are reserved for the captioning task.

The contrastive task operates over a batch of $B$ functions and names; it consists of aligning \texttt{[CLS]} text tokens $\mathit{CLS}$ with corresponding contrastive tokens $\mathit{Co}$:
\[
\mathcal{L}_{\text{Cross-Entropy}}(x, y) = -\frac{1}{B} \sum_{i=1}^{B} \log \frac{\exp(x_{i}^{\mathsf{T}}\,y_{i}/\sigma)}{\sum_{j=1}^{B}\exp(x_{i}^{\mathsf{T}}\,y_{j}/\sigma)}
\]
\[
\mathcal{L}_{\mathrm{Contrastive}} = \mathcal{L}_{\text{Cross-Entropy}}(\mathit{CLS},\mathit{Co}) 
+ \mathcal{L}_{\text{Cross-Entropy}}(\mathit{Co},\mathit{CLS})
\]

Here, $i$ and $j$ are indices for functions and names in the $i$-th and $j$-th pairs, while $\sigma$ is the temperature used to scale the logits. $L_{Contrastive}$ is the sum of function-to-name cross-entropy and name-to-function cross-entropy. This approach minimizes the distance between $x_{i}$ and $y_{i}$ while maximizing the distance between $x_{i}$ and $y_{j}$. Such alignment ensures that the function and text tokens in each pair remain in the same latent space.

The captioning loss uses function tokens, excluding the contrastive one, and unimodal text tokens, excluding \texttt{[CLS]}.

As shown in \autoref{fig:CoCa}, we first connect the multimodal text decoder to the unimodal text encoder, and then link the multimodal text decoder to the function encoder via cross-attention layers. As a result, the multimodal text decoder can leverage information from both the unimodal text encoder and the function encoder. Thus, during backpropagation, the weights of both encoders are updated accordingly. The output of the multimodal decoder is function names, as word sequences, and the loss function is the standard cross-entropy loss for generation tasks. 

Given ground truth words $\mathit{Name}$ and multimodal text tokens $\mathit{MMTTokens}$, the caption loss is as follows:
\[
\mathcal{L}_{\mathrm{Caption}} = -\sum_{t=1}^{n} \log \mathcal{P}_{\theta}(\mathit{Name}_{t} \mid \mathit{Name}_{<t}, \mathit{MMTTokens})
\]
Here, $\mathit{Name}_{t}$ is the ground truth word at time step $t$ and $\mathit{Name}_{<t}$ denotes previous words up to time step $t-1$. $\mathcal{P}_{\theta}(\mathit{Name}_{t} \mid \mathit{Name}_{<t}, MMTTokens)$ is the predicted probability of $\mathit{Name}_{t}$ given $\mathit{Name}_{<t}$ and $\mathit{MMTTokens}$.

The full
loss function of the training process is as follows:
\[
\mathcal{L}_{\mathrm{Full}} = \mathcal{L}_{\mathrm{Contrastive}} + \mathcal{L}_{\mathrm{Caption}}
\]

After  \textsc{Combo}, the function encoder can project function tokens into a latent space closely aligned with that generated by the unimodal text encoder. This alignment opens possibilities for zero-shot function-text retrieval~\cite{WangISSTA24}.

The multimodal text decoder can already generate imprecise function names.
However, pre-trained models perform best when fine-tuned only on their downstream task, which is why CoCa is fine-tuned on only the captioning task~\cite{coca}.
Consequently, we fine-tune the pre-trained model on function name prediction inspired by the encoder-decoder structure in GIT~\cite{git}. Along the way, we propose a new decoder, \textit{\textsc{Lord}}, that is more appropriate for function name prediction.

\section{\textsc{Lord}\label{sec:lord}}

An important challenge in function name prediction is achieving an optimal balance between recall and precision.
Specifically, it is crucial to prevent overwhelming a reverse engineer with excessive false positives, which are time-consuming to verify.
Preliminary work~\cite{ye2023mitigating}
indicates that traditional transformer approaches tend to be overconfident, prioritizing recall excessively. This issue is common to modern neural networks, especially in the face of distribution shifts~\cite{wei2022mitigating,kristiadi2020being}.
This bias is acceptable in problems such as translation and image captioning, where errors are easily noticeable and distribution shifts are small.
However, in binary analysis, a transformer should maintain precision by being more cautious.

We propose \textsc{Lord}, a new decoder and inference scheme parameterized by a confidence threshold. The embedding layer of this decoder is initialized with weights from the unimodal text encoder. To deepen semantic understanding, we introduce a new Masked Language Modeling (MLM) task (detailed in \autoref{sec:mbl}), inspired by BERT~\cite{devlin-bert}, to replace left-to-right teacher forcing.
During inference, we use a more precise autoregressive process referred to as flexible autoregression. It selects the word with the highest confidence score at each step until the score falls below the threshold. Details are provided in~\autoref{sec:smi}.

\begin{figure}[t]
    \centering
    \includegraphics[width=.9\linewidth]{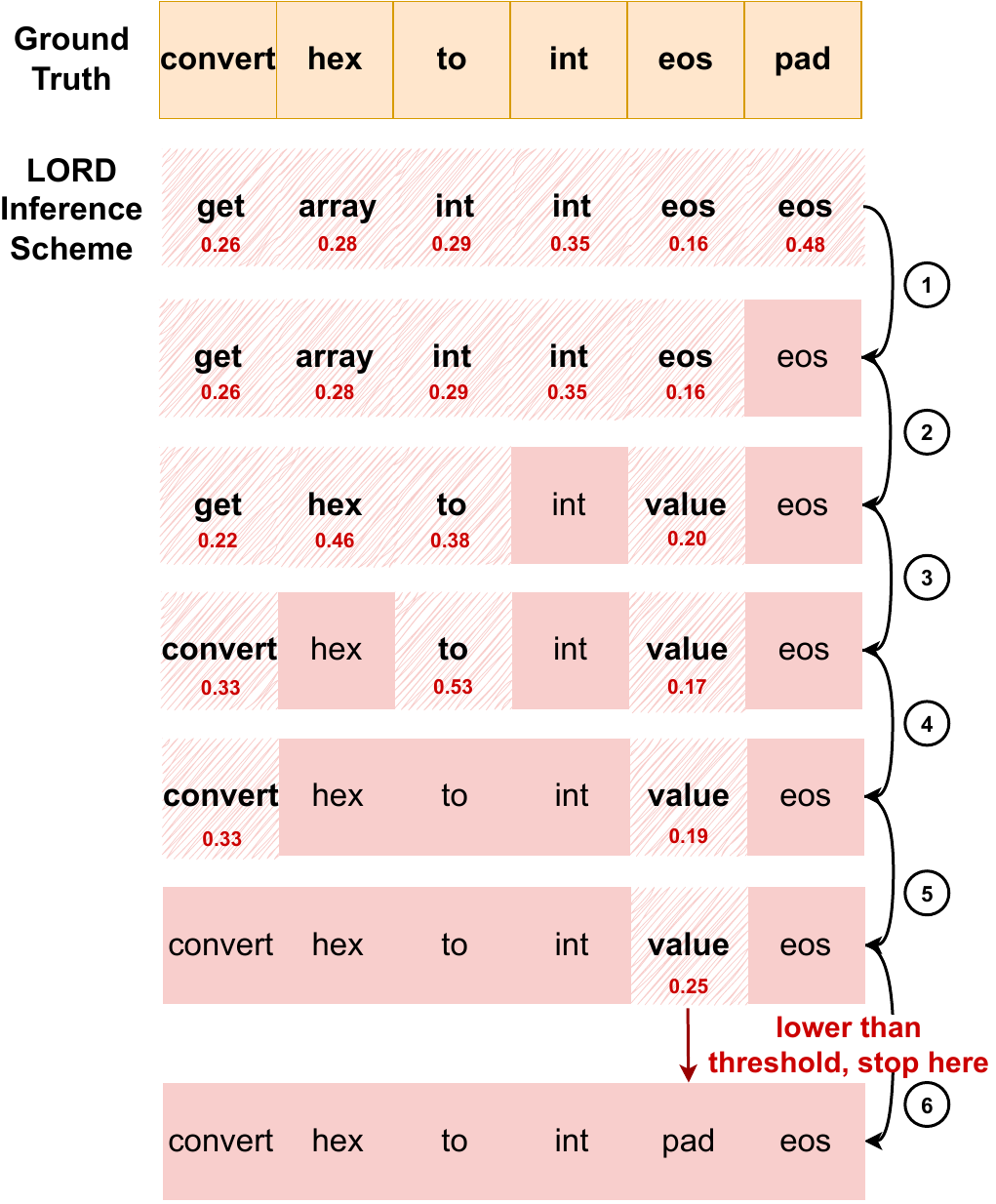}
    \caption{A toy example of the flexible autoregressive process.}
    \label{fig:flexibledecoder}
\end{figure}

\subsection{Masked Language Modeling}\label{sec:mbl}

During training, a traditional generative transformer learns through teacher forcing~\cite{kolen01book}. Each word is predicted with access to the preceding words as context. During inference, autoregression predicts words sequentially from left to right until the \texttt{[EOS]} word is reached. This approach is valid because, with enough transformer blocks, autoregression can express any sequence distribution. However, teacher forcing simplifies the prediction of later words compared to the first word and can fail planning tasks~\cite{bachmann24pitfallsnexttokenprediction}.

We propose a new MLM task to replace teacher forcing. We train the decoder to predict masked words given unmasked ones. This allows for more possibilities of masking and helps to mitigate the issue of short function names. Additionally, this lets the decoder learn deeper insights into the relationship between function names and binary code.

In line with the traditional transformer, where the context (previous words) never includes \texttt{[EOS]} and \texttt{[PAD]} words, we consistently mask these words. %
However, traditional MLM tasks randomly mask 15\% of words within a long text corpus~\cite{wettig2022should}, yet
function names are typically short, often consisting of three to five words. A 15\% masking rate results in a task of unmasking too few tokens. Consequently, we need to mask a higher proportion of words.
We follow the intuition that the number of words and the number of masked words should be related. While we want to focus on challenging cases, we must also create contexts containing nearly all words since such context will occur during inference. For a function name with $n$ words, we define the number of masked words, from 0 to $n$, as a random variable $M$, which follows the discrete probability distribution $\mathit{Softmax}(C)$, where $C$ is a vector of dimension $n+1$ with $C_i = 1 + i/n$. In our MLM task, $M$ words are thus masked randomly among all possible subsets.

For instance, consider the four-word function name \texttt{convert\_hex\_to\_int}. The number of masked words $M$ follows the distribution {\small $\mathit{Softmax}([1,1.25,1.5,1.75,2]) \approx [0.114,0.147,0.188,0.241,0.310]$}.
The formula ensures that masking all words is the most probable event but maintains a decent probability of masking a few words. If two words happen to be masked, then a subtask could be to recover  \texttt{convert\_hex\_to\_int} from \texttt{[M]\_hex\_[M]\_int} given function tokens described in~\autoref{sec:cc} as extra context.

\subsection{Flexible Autoregression}\label{sec:smi}

\textsc{Lord} is not limited to left-to-right inference.
It could decode the function name in one step~\cite{qi20prophetnet,monea2023pass}.
However, decoding the whole name in one step introduces  difficulties in inferring relations between words, as the decoder better approximates the conditional distribution of a word given  context~\cite{santilli2023accelerating}. %
As the example shows in~\autoref{fig:flexibledecoder}, the decoder might initially predict \texttt{int} for both positions 2 and 3. However, once \texttt{int} is placed at position 3, the likelihood of \texttt{int} also being at position 2 becomes very low.

Hence, we introduce a new autoregressive process called the flexible autoregression.
It starts with an initial prediction containing only masked words.
At each inference step, it runs our decoder on the current prediction.\footnote{Because \texttt{[EOS]} words are never part of the context during our MLM task, we have to mask them from the decoder.}
The decoder estimates the possible words at each masked position. The autoregression picks the word \( w \) with the overall highest confidence score (probability) at any given position.
It stops if that confidence score is smaller than the confidence threshold \( T \) or if all words are unmasked. Otherwise, it replaces the corresponding masked word with the word $w$, and it continues. 
The confidence threshold \( T \) is a parameter to flexible autoregression for addressing distribution shifts between  the training and test sets, thereby improving the precision. We select the threshold that achieves the best \( F_1 \) score on the validation set.

We give a toy example of the process in~\autoref{fig:flexibledecoder}. 
It first infers the \texttt{[EOS]} word; as explained before this special word is masked from the decoder context.  It continues by positioning the \texttt{int} word in the fourth position because the confidence score at the third position is lower ($0.29$ vs $0.35$), leaving some positions for earlier words. Then, it notices an array that contains, in fact, a hexadecimal number, so it adds the \texttt{hex} word at the second position. It concludes correctly that the function is a conversion from a hexadecimal number to an integer by correctly placing the preposition \texttt{to} and the verb \texttt{convert} in two steps. 

\section{Evaluation}

In this section, we evaluate \ProjectName comprehensively.
In~\autoref{sec:methodology}, we describe the evaluation methodology, including the dataset, metrics for function name prediction, and competitors.
With this methodology, we evaluate \ProjectName in the cross-binary setting (\autoref{sec:crossbinarysetting}) and the cross-project setting (\autoref{sec:crossprojectsetting}). In both settings, we investigate whether \ProjectName outperforms competitors and analyze \ProjectName predictions. Furthermore, we introduce a strict setting that removes shared components across projects to evaluate generalization capabilities~(\mbox{\autoref{sec:strict_setting})}.
We perform an ablation study on \ProjectName to validate our contributions against other design choices (\autoref{sec:ablation}).
The discussion in~\autoref{sec:discussion} indicates that, in the challenging cross-project setting, \ProjectName's predicted function names can be meaningful even if different from the ground truth.

\subsection{Methodology\label{sec:methodology}}

\myparagraph{Dataset} We use the Punstrip~\cite{acsac20-punstrip} dataset, as does XFL~\cite{oakland23-xfl}. It contains 741,724 functions originating from 10,047 C binaries extracted from more than three thousand Debian packages\footnote{For simplicity, we refer to packages as projects in this paper.}, which have been pre-compiled with different compilers and compiler versions. Following previous work~\cite{acsac20-punstrip,oakland23-xfl}, we exclude empty functions, overlapping functions, and locally bound symbols, and we use the ELF symbol table to obtain function names.

\myparagraph{Tokenizer} To predict words in function names, we first have to transform function names into sequences of words.
We use the tokenizer by Patrick-Evans et al.~\cite{oakland23-xfl}, which assembles a vocabulary of 1024 words, along with a set of words for each function name. We wrote a recursive version of the tokenizer algorithm to produce sequences instead of sets.
\myparagraph{Settings} We adopt the two settings defined by Xiong et al.\mbox{~\cite{XiongASE23}}: the cross-project setting and the cross-binary setting, to understand both practical performance and the model's generalization capability. In both settings, we split functions into training, validation, and test sets, which comprise approximately 80\%, 10\%, and 10\% of the total dataset, respectively. %
In the {\bf cross-project} setting, an entire project is allocated to one of the sets, ensuring that functions from the same project belong to the same set. In the {\bf cross-binary} setting, we only ensure that functions from the same binary file belong to the same set. 
The cross-project setting is more realistic since, in most use cases, a reverse engineer is working on an unseen project.
However, even different projects may reuse code from certain APIs, design patterns, and templates, which is also expected in a real-world deployment of function name prediction. 
To specifically evaluate generalization abilities, we define the \mbox{\bf strict setting} on top of the cross-project setting, which removes potential shared components across projects.
\myparagraph{Pre-processing}
As explained in~\autoref{sec:cc}, \ProjectName employs two function embeddings, \textsc{Dexter}~\cite{oakland23-xfl} and \textsc{\textsc{Clap}}~\cite{WangISSTA24}, plus one basic block embedding, \textsc{PalmTree}~\cite{palmtree}. All embeddings employ their own pre-processing, tied to specific disassemblers.
\textsc{Dexter} uses the radare2~\cite{radare2book} (v2-5.5.4) disassembler. Because it relies heavily on vocabularies extracted from the training set (e.g., external calls, operations), we had to train distinct \textsc{Dexter} models for the cross-binary and cross-project settings. Note that \textsc{Dexter} reads function boundaries from the symbol table because it considers recovering function boundaries from stripped binary code an orthogonal task.
To obtain \textsc{\textsc{Clap}} and \textsc{PalmTree} embeddings, we employ pre-trained models from their authors. %
We use IDA Pro (v7.6) as a disassembler to obtain \textsc{\textsc{Clap}} function embeddings. Likewise, we compute \textsc{PalmTree} instruction embeddings with angr~\cite{angr} (v9.2.93) as a disassembler. %
We average the instruction embeddings of each basic block to obtain sequences of basic block embeddings for each function.%

\myparagraph{Metrics\label{sec:metrics}} Let us define the list of predicted function names \(\hat{X}_1, \ldots, \hat{X}_B\) and the list of ground truth function names \(X_1, \ldots, X_B\).
Metrics commonly used for function name prediction consider \(\hat{X}_i\) as a set of words \(\{\hat{x}_1, \ldots, \hat{x}_j \}\), while the ground truth $X_i$ is another set  $\{x_1, \ldots, x_k\}$.
True positives for the $i$ th function ($\TP_i$), false positives ($\FP_i$), and false negatives ($\FN_i$) are defined as:
\[ \TP_i = |\{\hat{x}_i\} \cap \{x_i\}|, \FP_i = |\{\hat{x}_i\} \setminus \{x_i\}|, \FN_i = |\{x_i\} \setminus \{\hat{x}_i\}| \]

Because it gives equal weight to each word, we adopt the micro-average definition of the precision $P$ and recall $R$: 
\[
P = \frac{\sum_{i=1}^{N} \TP_i}{\sum_{i=1}^{N} (\TP_i + \FP_i)}, \quad
R = \frac{\sum_{i=1}^{N} \TP_i}{\sum_{i=1}^{N}(\TP_i + \FN_i)}
\]

Recall is fundamental to offer a decent number of predictions.
Moreover, precision is equally crucial in our context, since false positives can only be discovered after complex human analysis of assembly code.
Hence, a good balance between recall and precision is necessary for function name prediction. The $F_1$ score measures this balance by the harmonic mean of precision and recall.%

\myparagraph{Word order metrics} Nevertheless, $F_1$ does not consider the order of words or any repetitions of words in the ground truth. Therefore, we also consider two classical metrics for image captioning and translation that operate on n-grams.
The first metric, ROUGE-L~\cite{lin-och04acl}, referred to as RougeL in this paper, measures the longest common subsequences and thus places a greater emphasis on recall. The second metric, the smoothed version of BLEU-4~\cite{lin-och04coling}, referred to as Bleu in this paper, focuses on n-gram precision scores. Despite the application of a brevity penalty, in the end, it remains a precision metric. Therefore, although these metrics are crucial for capturing word order, $F_1$ still remains the primary metric of function name prediction.

\myparagraph{Embedding metrics} 
As an alternative to classical metrics, embedding-based metrics, such as \textsc{BERTScore}\mbox{~\cite{bertscore}} and VarCLR\mbox{~\cite{chen2022varclr}}, promise to measure semantic similarity between predictions and ground truth with the distance between their embeddings.
In our evaluation, we include measurements of VarCLR, as it is trained on software and designed for variable names.
However, it comes with inherent limitations (at times, counter-intuitive judgments and issues with empty predictions; see\mbox{~\autoref{sec:discussion})}, which is why we emphasize the need for $F_1$ as a robust performance metric.
\myparagraph{Free functions\label{par:freenames}}  To accurately simulate a real-world scenario, twenty functions that can be inferred automatically by static analysis (e.g., \texttt{csu\_init}) are treated as perfectly predicted in the cross-binary and cross-project settings. This function list is adopted from Patrick-Evans et al.'s evaluation of XFL\mbox{~\cite{xfl-github}} and reported in\mbox{~\autoref{appendix:freebies}}.
\myparagraph{Implementation\label{sec:implementation}} \ProjectName{} is implemented in Pytorch and utilizes parts of CoCa and GIT implementations. %
All experiments were run in two weeks on a server with access to one terabyte of main memory and four NVIDIA H100 GPUs, each with 80GB of VRAM.
The \ProjectName transformer architecture features a token dimension $d$ of 768 and a batch size of 512. The number of function patches $k_1$ and image tokens $k_2$ are 82 and 64, respectively.  Function names consist of up to 20 words.  \ProjectName transformer blocks use multi-head attention layers with 32 heads of dimension 24. The large number of heads is effective for handling multiple embeddings, while the relatively low dimension helps avoid overfitting. The pre-trained model contains 153 million learnable parameters, decreasing to 138 million following fine-tuning. Further implementation details are documented in our artifact~\cite{blens-artifact}.

\myparagraph{Training} Both the pre-training and fine-tuning phases last 200 epochs. Every 10 epochs during fine-tuning, confidence thresholds are evaluated to find the optimal threshold according to the $F_1$ score on the validation set. The model is then saved along with its threshold. In the ablation study, each phase has 80 epochs, and we obtain a confidence threshold as well as a model every 4 epochs. Eventually, the best model on the validation set is evaluated over the test set using the corresponding confidence threshold. The confidence threshold turned out to be 0.194 for the cross-project setting and 0.398 for the cross-binary setting.

\myparagraph{Competitors} We consider three main competitors for the function name prediction task.
AsmDepictor by Kim et al.~\cite{kim23asiaccs} translates instruction sequences to function names with a transformer.
SymLM by Jin et al.~\cite{jin22ccs} is a transformer that considers both the target function and the function calling context. External calls are embedded, while internal functions are represented by Trex embeddings~\cite{pei21arxiv}.
XFL by Patrick-Evans et al.~\cite{oakland23-xfl} predicts word sets with PFastreXML~\cite{JainPV16} and \textsc{Dexter} before ordering them with an n-gram model.

\myparagraph{Adaptation} Our competitors use different function name tokenizers; thus, we adapted each to work with our tokenizer and trained them in both the cross-binary and cross-project settings for a fair comparison.
With their pre-processing phase, \ProjectName and XFL gathered 436,941 functions, while SymLM and AsmDepictor gathered 232,393 and 318,372 functions, respectively. Due to these differences, we had to train specific versions of \ProjectName on SymLM and AsmDepictor datasets for both the cross-binary and cross-project settings. We call the first model \textsc{BL-s} and the second \textsc{BL-a}.

\myparagraph{HexT5} HexT5~\cite{XiongASE23} is also an interesting competitor; however, the source code for the learning phase and data processing is not available. Therefore, we could only reimplement the binary processing step based on the provided input examples and use the pre-trained model. This model uses a vastly different tokenizer. Consequently, we conduct a qualitative evaluation of HexT5, which is discussed further in~\autoref{sec:discussion}.

\begin{table}
    \centering
    \begin{tabularx}{\linewidth}{Xcccc}
        \toprule
        Model & $F_1$ & RougeL & Bleu & VarCLR \\
        \midrule
        {\bf \ProjectName}        & {\bf 0.772} & \bf 0.699 & {\bf 0.582} & \bf 0.819 \\
        XFL          & 0.625 & 0.527 & 0.325 & 0.720 \\
        \midrule
        {\bf \textsc{BL-s}}  & {\bf 0.789} & \bf 0.689 & \bf 0.652 & \bf  0.817 \\
        SymLM      & 0.704 & 0.472 & 0.107 & 0.736 \\
        \midrule
        {\bf \textsc{BL-a}}  & {\bf 0.763} & \bf 0.693 & {\bf 0.596} & \bf 0.819 \\
        AsmDep. & 0.407 & 0.432 & 0.279 & 0.649 \\
        \bottomrule
    \end{tabularx}
    \medskip

    \caption{\label{fig:binary:testScores}
Results in the cross-binary setting. \textsc{BL-a} and \textsc{BL-s} are variants trained on functions obtained from SymLM and AsmDepictor pre-processing.} 

\end{table}

\begin{table}[t]
    \centering
    \begin{tabularx}{\linewidth}{Xcccc}
    \toprule
    Word & Occurrences & Prec. & Recall & $F_1$ \\
    \midrule
    \texttt{ocaml} & 1852 & 0.983 & 0.990 & 0.987 \\
    {\bf \texttt{get}} & {\bf 1350} & {\bf 0.864} & {\bf 0.740} & {\bf 0.797} \\
    {\bf \texttt{set}} & {\bf 687} & {\bf 0.880} & {\bf 0.702} & {\bf 0.781} \\
    {\bf \texttt{string}} & {\bf 683} & {\bf 0.898} & {\bf 0.783} & {\bf 0.836} \\
    \texttt{2} & 550 & 0.815 & 0.867 & 0.840 \\
    \texttt{4} & 418 & 0.995 & 0.965 & 0.980 \\
    \texttt{fun} & 403 & 0.681 & 0.492 & 0.571 \\
    \texttt{parse} & 378 & 0.902 & 0.595 & 0.717 \\
    \texttt{to} & 365 & 0.868 & 0.660 & 0.750 \\
    {\bf \texttt{read}} & {\bf 362} & {\bf 0.881} & {\bf 0.709} & {\bf 0.786} \\
    {\bf \texttt{print}} & {\bf 361} & {\bf 0.842} & {\bf 0.738} &{\bf  0.787} \\
    \texttt{initialise} & 360 &  0.806 &  0.613 & 0.696 \\
    \texttt{reiser} & 351 & 0.997 & 0.989 & 0.993 \\
    {\bf \texttt{list}} & {\bf 340} & {\bf  0.888} & {\bf 0.572} & {\bf 0.696} \\
    {\bf \texttt{free}} & {\bf 326} & {\bf 0.920} & {\bf 0.837} & {\bf 0.876} \\
    \texttt{is} & 317 & 0.840 & 0.698 & 0.762 \\
    {\bf \texttt{buffer}} & {\bf 314} & {\bf0.932} & {\bf 0.722} & {\bf 0.814} \\
    \texttt{lwt} & 273 & 0.996 & 0.424 & 0.595 \\
    \texttt{name} & 252 & 0.896 & 0.677 & 0.771 \\
     {\bf \texttt{integer}} &  {\bf 250} &  {\bf 0.887} &  {\bf 0.779} &  {\bf 0.830} \\
    \bottomrule
    \end{tabularx}

    \caption{\label{tab:binary:tokens}
Top 20 predicted words in the cross-binary setting. Functions given for free (described in~\autoref{par:freenames}) are discarded.}

\end{table}

\subsection{Cross-Binary Setting\label{sec:crossbinarysetting}}

Results in the cross-binary setting are summarized in~\autoref{fig:binary:testScores}.
\ProjectName outperforms XFL by 23.5\%, 32.6\%, 79.1\%, and 13.7\% in terms of  $F_1$, RougeL, Bleu, and VarCLR scores, respectively.
\textsc{BL-s} outperforms SymLM by 12\%, 45.9\%, 507\%, and 11.1\% in terms of  $F_1$, RougeL, Bleu, and VarCLR scores, respectively.
\textsc{BL-a} outperforms AsmDepictor by 87.7\%, 60.5\%, 113\%, and 26.1\% in terms of  $F_1$, RougeL, Bleu, and VarCLR scores, respectively.
\ProjectName obtains significantly better metrics, especially RougeL and Bleu scores.
Moreover, \ProjectName achieves an $F_1$ score of $0.772$ due to a high precision of $0.917$ and a good recall of $0.666$. \ProjectName RougeL, Bleu, and VarCLR scores are $0.699$, $0.582$, and $0.819$, which suggest a good capacity to recover subtle grammar details and produce meaningful names.

In~\autoref{tab:binary:tokens}, we report the most predicted words of \ProjectName in the cross-binary setting, along with their precisions, recalls, and $F_1$ scores. Note that functions given for free (described in~\autoref{par:freenames}) are discarded from scores.
Note further that \ProjectName is nearly perfect at predicting words specific to \texttt{ocaml} and certain libraries, such as \texttt{reiser}. These words occur as prefixes in many functions and are straightforward to recover: OCaml code has distinct patterns, and ReiserFS-specific functions are prefixed by \texttt{reiser4}, which is part of the reason we also achieve a very high score on the word \texttt{4}. Moreover, some standard OCaml functions are present in multiple binaries, and ReiserFS functions are shared across three binaries (prefixes and code sharing are specifically addressed in \autoref{sec:strict_setting}).

\ProjectName also achieves good results on other words.
For instance, on \texttt{get} and \texttt{set}, which are common general-purpose terms, \ProjectName attains $F_1$ scores of $0.797$ and $0.781$, respectively.
On words related to strings and input/output operations, such as \texttt{string}, \texttt{parse}, \texttt{read}, and \texttt{print}, \ProjectName achieves $F_1$ scores of $0.836$, $0.717$, $0.786$, and $0.787$, respectively. 
Finally, on words related to low-level operations such as \texttt{initialise}, \texttt{free}, \texttt{list}, and \texttt{buffer}, \ProjectName achieves $F_1$ scores of $0.696$, $0.876$, $0.696$, and $0.814$, respectively.

\myparagraph{Conclusion}
In the cross-binary setting, \ProjectName achieves a notable $F_1$ score of 0.771, with a critical precision of 0.917. Thanks to our contributions, \ProjectName captures word order, domain-specific words, and general-purpose words, improving the state of the art by 12\% on $F_1$, 32.6\% on RougeL,  79.1\%  on Bleu, and 11.1\% on VarCLR scores.

\begin{table}[t]
    \centering
    \begin{tabularx}{\linewidth}{Xcccc}
        \toprule
        Model & $F_1$ & RougeL & Bleu & VarCLR \\
        \midrule
        {\bf \ProjectName} & {\bf 0.460} & {\bf 0.393} & {\bf 0.242} & \bf 0.648\\ 
        XFL & 0.295 & 0.222 & 0.025 & 0.560\\
        \midrule
        {\bf \textsc{BL-s}} & {\bf 0.394} & {\bf 0.294} & {\bf 0.221} & \bf 0.599\\
        SymLM & 0.277 & 0.171 & 0.023 & 0.534\\
        \midrule
        {\bf \textsc{BL-a}} & {\bf 0.455} & {\bf 0.399} & {\bf 0.258} &  \bf 0.652 \\
        AsmDep. & 0.200 & 0.220 & 0.089 &  0.520 \\
        \bottomrule
    \end{tabularx}
\caption{\label{fig:project:testScores}Results in the cross-project setting. } 

\end{table}

\begin{table}[t]
\centering
\begin{tabularx}{\linewidth}{X|rr|rc}
\toprule
\multicolumn{1}{r|}{}  & \multicolumn{2}{c|}{Cross-Project} & \multicolumn{2}{c}{Strict} \\
Word & {Occ.} & {$F_1$} & {Occ.} & {$F_1$} \\     
\midrule
\texttt{ocaml} & 2162 & 0.975 & 0 &  -\\
\texttt{get} & 1132 & 0.365 & 813 & 0.313\\
{\bf \texttt{string}} & {\bf 1050} & {\bf 0.498} & \bf 817 &  \bf 0.465 \\
{\bf \texttt{free}} & {\bf 519} & {\bf 0.750} & \bf 379 & \bf 0.688 \\
{\bf \texttt{type}} & {\bf 465} & {\bf 0.780} & \bf 150 & \bf 0.599 \\
\texttt{initialise} & 435 & 0.352 & 372 & 0.305 \\
{\bf \texttt{fun}} & {\bf 401} & {\bf 0.631} & \bf 365 & \bf 0.640 \\
\texttt{set} & 392 & 0.336 & 307 & 0.295 \\
\texttt{visit} & 389 & 0.960 & 0 & - \\
\texttt{soap} & 337 & 0.998 & 0 & - \\
{\bf \texttt{print}} & {\bf 332} & {\bf 0.486} & \bf 290 & \bf 0.434 \\
\texttt{read} & 326 & 0.418 & 311 & 0.441 \\
\texttt{2} & 305 & 0.212 & 207 & 0.110 \\
\texttt{curry} & 304 & 1.000 & 0 & - \\
\texttt{path} & 288 & 0.368 & 279 & 0.378 \\
\texttt{name} & 285 & 0.417 & 255 & 0.388 \\
\texttt{usal} & 266 & 0.981 & 0 & - \\
{\bf \texttt{error}} & {\bf 263} & {\bf 0.675} & \bf 208 & \bf 0.601 \\
{\bf \texttt{open}} & {\bf 254} & {\bf 0.556} & \bf 238 & \bf 0.584 \\
\texttt{information} & 244  & 0.650 & 58 & 0.205 \\
\bottomrule
\end{tabularx}

    \caption{Top 20 words predicted in the cross-project setting with their occurrences and $F_1$ scores in the cross-project and strict settings.\label{tab:project:tokens} }
    
\end{table}

\subsection{Cross-Project Setting\label{sec:crossprojectsetting}}

Results in the cross-project setting are summarized in~\autoref{fig:project:testScores}.
\ProjectName outperforms XFL by 55.7\%, 76.7\%, 887\%, and 15.9\% in terms of  $F_1$, RougeL, Bleu, and VarCLR scores, respectively.
\textsc{BL-s} outperforms SymLM by 42.2\%, 71.7\%, 855\%, and 12.1\% in terms of  $F_1$, RougeL, Bleu, and VarCLR scores, respectively.
\textsc{BL-a} outperforms AsmDepictor by 128\%, 81.1\%, 188\%, and 25.3\% in terms of  $F_1$, RougeL, Bleu, and VarCLR scores, respectively.

Again, \ProjectName achieves significantly better metrics, especially $F_1$, RougeL, and Bleu scores.
Moreover, \ProjectName achieves an $F_1$ of $0.460$ due to a good precision of $0.655$ and a decent recall of $0.354$. \ProjectName RougeL, Bleu, and VarCLR scores are $0.393$, $0.242$, and $0.648$ which suggest a decent capacity to recover word order and produce meaningful function names. 

In~\autoref{tab:project:tokens}, we report the most predicted words of \ProjectName in the cross-project setting, along with their $F_1$ scores. Again, functions given for free (described in~\autoref{par:freenames}) are discarded.
Even in the cross-project setting, the extremely high scores on some words (e.g., an $F_1$ of 0.960 for the word \texttt{visit}) can be attributed to the training set being highly relevant to a new target program.
\ProjectName obtains nearly perfect scores for the word \texttt{ocaml} because our dataset includes approximately 80 OCaml libraries as projects. Additionally, the source code of the OCaml standard library is shared across three projects, while individual OCaml libraries employ this code base, highlighting interconnections between projects and shared code components. Likewise,  the word \texttt{curry} is frequently used by the OCaml compiler.
Similarly, \ProjectName achieves impressive metrics for the word \texttt{visit}. This word appears in both the test and training sets due to the common reliance on the QAPI interface. The QAPI (QEMU API) schema defines a set of data types explored with the visitor pattern.
For the word \texttt{soap}, the project gridsite-clients in the test set uses the SOAP (Simple Object Access Protocol) for web service communications, a pattern similarly observed in the training data from the parent project gridsite and the related project lfc. %
Lastly, \ProjectName can accurately predict the word \texttt{usal} thanks to functions from the libusal library inside projects like genisoimage and wodim. The libusal library supports creating CD and DVD images.

\pagebreak

We can categorize the rest of the words into different operation groups.
Firstly, words related to data manipulation include \texttt{get}, \texttt{set}, \texttt{read}, \texttt{name}, and \texttt{path}. \ProjectName achieves $F_1$ scores of 0.365, 0.336, 0.418, 0.417, and 0.368, respectively.
Secondly, for memory management, words include \texttt{free}, \texttt{initialise}, and \texttt{open}. \ProjectName exhibits good performance with $F_1$ scores of 0.750, 0.352, and 0.556, respectively. Notably, \texttt{free} shows a high $F_1$ score, demonstrating BLens's ability to identify memory liberation.
Thirdly, on words related to output and notification, such as \texttt{print} and \texttt{error}, \ProjectName obtains $F_1$ scores of 0.486 and 0.675. The higher $F_1$ score for \texttt{error} indicates good performance in identifying error handling code.
Lastly, on miscellaneous words such as \texttt{string}, \texttt{type}, \texttt{fun}, and \texttt{information}, \ProjectName obtains $F_1$ scores of 0.498, 0.780, 0.631, and 0.650, respectively.
\myparagraph{Conclusion}
In the cross-project setting, \ProjectName significantly outperforms SotA methods, particularly on metrics capturing word order, with 42.2\% $F_1$, 71.7\% RougeL, 188\% Bleu, and 12.1\% VarCLR score improvement.
\ProjectName achieves notable $F_1$ scores on some general-purpose words, for instance, 0.750 for \texttt{free} and 0.675 for \texttt{error}.
While $0.460$ $F_1$, $0.393$ RougeL, and $0.24$ Bleu scores seem low in absolute terms, predictions by BLens may be meaningful even when different from the ground truth, as we discuss in~\autoref{sec:discussion}.
\begin{table}
    \centering
    \begin{tabularx}{\linewidth}{Xcccc}
        \toprule
        Model & $F_1$ & RougeL & Bleu & VarCLR \\
        \midrule
        \bf \ProjectName & \bf 0.294 & \bf 0.243 & \bf 0.094 & \bf 0.568\\ 
        XFL & 0.085 & 0.055 & 0.001 &0.474\\
        \midrule
        \bf \textsc{BL-s} & \bf 0.299 & \bf 0.240 & \bf 0.110 & \bf 0.571\\
        SymLM & 0.195 & 0.134 & 0.011 & 0.518\\
        \midrule
        \bf \textsc{BL-a} & \bf 0.321 & \bf 0.272 & \bf 0.098 &  \bf 0.585 \\
        AsmDep. & 0.090 & 0.085 & 0.034 &  0.445 \\
        \bottomrule
    \end{tabularx}
    \caption{\label{fig:project:challenge}
Results in the strict setting.} 
\end{table}
\subsection{Strict Setting}\label{sec:strict_setting}
This experimental setting extends the cross-project setting to rigorously evaluate generalization capabilities by minimizing potential shared code components across projects. Ideally, this would involve deduplication at the source code level; however, as the dataset consists of pre-compiled binaries, we define a set of heuristics derived from manual analysis to identify and remove as many similar components from binaries as possible. %
Initially, we remove hash duplicates and, for completeness, also freely given functions (see \mbox{\autoref{par:freenames}}). We observe a relatively consistent drop of about 0.059 in the $F_1$ score across most methods. However, the impact is minimal on SymLM and BL-S, as SymLM's pre-processing has already removed part of these functions.

To identify shared code where the hash differs after compilation, we analyze the words in function names. Note that we cannot simply remove all functions from testing whose names occur in the training set, as this would exclude many benign cases such as \texttt{main} or \texttt{print\_help}. 
Shared code typically arises from statically-linked runtime functions and library calls, which often share name prefixes (e.g., \texttt{ocaml}, \texttt{soap}, \texttt{usal}), or non-prefix words that frequently co-occur with prefixes (e.g., \texttt{curry} co-occurs with \texttt{ocaml}, as it denotes curried functions). %
We generate an exclusion list of prefixes and co-occurring words found from a semi-automatic analysis of the top 200 performing words and binaries. Any function whose name appears in the training set, contains an excluded word, and has an $F_1 > 0$, is removed from the test set. Allowing functions with an $F_1$ of $0$ avoids removing some functions containing excluded words by chance. From the remaining function names in the test set, we remove all excluded words.

Ultimately, 40 words and 6491 functions (2644 freely given functions, 2526 hash duplicates, and 1321 other duplicates) are removed from 1024 words and 23,875 initial functions in the test set.
In\mbox{~\autoref{tab:project:tokens}}, we report the new occurrences and $F_1$ scores of the previously most predicted words in the strict setting. We observe that only general purpose words such as \texttt{string}, \texttt{free}, and \texttt{type} remain, with  $F_1$ scores of $0.465$, $0.688$, and $0.599$, respectively.
Results are summarized in\mbox{~\autoref{fig:project:challenge}}.
While BLens's $F_1$ score decreases from $0.460$ to $0.294$, it still outperforms all competitors.

\myparagraph{Conclusion}  In the strict setting, BLens continues to significantly outperform the state of the art, with 53.3\% $F_1$, 79.4\% RougeL, 188\% Bleu, and 10.3\% VarCLR improvement. The widening of the gap between BLens and other methods in terms of $F_1$ highlights BLens's generalization.

\subsection{Ablation Study\label{sec:ablation}}

We now evaluate the impact of each component of \ProjectName to show that they are all beneficial.
We start by evaluating the impact of \textsc{Combo}, which aligns multiple embeddings together in a joint space with function names during pre-training.
Then, we assess the contribution of each input embedding (\textsc{\textsc{Clap}}, \textsc{PalmTree}, and \textsc{Dexter}) by evaluating each subset of these embeddings.
Finally, we investigate the advantages of \textsc{Lord} over the traditional decoder architecture and the direct use of \textsc{Combo} as a decoder.
We conduct the ablation in the cross-project setting with training phases of 80 epochs, optimizing the confidence threshold every four epochs.

\begin{table}
    \centering
    
    \begin{tabularx}{\linewidth}{Xcccc}
        \toprule
        Model & $F_1$ & RougeL & Bleu & VarCLR\\
        \midrule
        {\bf \ProjectName} & {\bf 0.445} &   {\bf 0.376} & {\bf 0.207} & \bf 0.639\\
        \textsc{BL-np} & 0.287 &   0.257 & 0.057 & 0.554\\
        \bottomrule
    \end{tabularx}
    \caption{\textsc{Combo} ablation results. \textsc{BL-np} is a variant trained without a pre-training phase. Ablation models are trained for only 80 epochs.}
    \label{tab:ablation:noCoCa:testScores}
\end{table}

\begin{table}
    \centering
    
    \begin{tabularx}{\linewidth}{Xcccc}
        \toprule
        Ensemble & $F_1$ & RougeL & Bleu & VarCLR \\
        \midrule
        {\bf C+P+D} & {\bf 0.445} &   0.376 & {\bf 0.207} & \bf 0.639 \\
        C+D         & 0.438 &   {\bf 0.378} & 0.196  &  0.638 \\
        C+P         & 0.425 &   0.370 & 0.182  & 0.629\\
        P+D         & 0.364 &   0.293 & 0.096  & 0.595\\
        C           & 0.425 &   0.366 & 0.184  & 0.633\\
        P           & 0.352 &   0.293 & 0.109  & 0.593\\
        D           & 0.310 &   0.254 & 0.060  & 0.574\\
        \bottomrule
    \end{tabularx}

    \caption{Input embeddings ablation results. C: \textsc{\textsc{Clap}}, P: \textsc{PalmTree}, D: \textsc{Dexter}, +: Combination of embeddings. 
    \label{tab:ablation:embeddings:testScores}    }
\end{table}

\subsubsection{\textsc{Combo}}\label{sec:nococa}

We define a new model, \textsc{BL-np}, which is trained without the \textsc{Combo} pre-training phase. This model starts with the ensemble encoder and turns embeddings into patches. Then, instead of using the function encoder pre-trained on two tasks, this model simply passes these patches to  \textsc{Lord} function name decoder for fine-tuning.
In~\autoref{tab:ablation:noCoCa:testScores}, we report metrics of \textsc{BL-np} and the base model \ProjectName. %
\ProjectName significantly outperforms \textsc{BL-np} across all metrics, with  $F_1$ and VarCLR  scores of  0.445 and 0.639, compared to \textsc{BL-np}  0.287  and 0.554, respectively. This higher performance is further reflected in the RougeL and Bleu scores, which are 0.376 and 0.207 for \ProjectName, compared to 0.257 and 0.057 for \textsc{BL-np}, respectively.

\myparagraph{Conclusion}
The $F_1$ score gain of $0.158$ clearly demonstrates the relevance of the contrastive captioning task.

\begin{figure}[t]
    \includegraphics[width=1\linewidth,trim=7 7 6 5,clip]{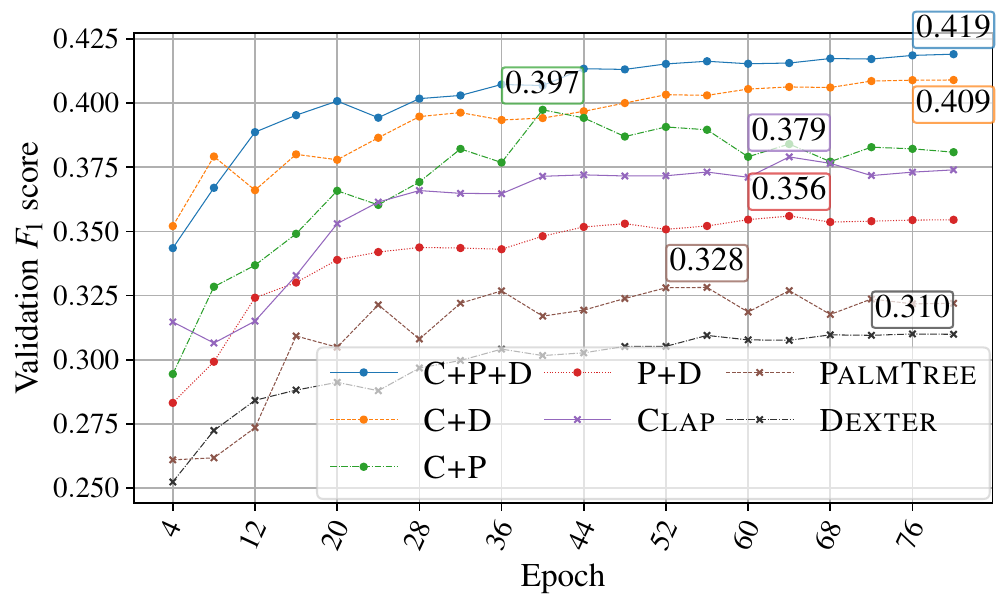}
     \caption{Curve of validation $F_1$ scores over the fine-tuning for the input embeddings ablation models.}
   \label{fig:embeddings:validationF1}
\end{figure}

\subsubsection{Input Embeddings}

Thanks to  \textsc{Combo}, \ProjectName incorporates multiple embeddings. The original model used \textsc{\textsc{Clap}}, \textsc{PalmTree}, and \textsc{Dexter}. Now, we train models with all possible combinations of these embeddings.
We report the metrics of each model in~\autoref{tab:ablation:embeddings:testScores}. We remark that \textsc{\textsc{Clap}} is the most effective embedding, evidenced by its $F_1$ score of 0.425, compared to \textsc{PalmTree} and \textsc{Dexter}, which have $F_1$ scores of 0.352 and 0.310, respectively.
Combining all three embeddings achieves the best $F_1$ score of $0.445$. Although the improvement over the combination of \textsc{Clap} and \textsc{Dexter} amounts to merely $0.007$ $F_1$ score, an analysis in~\autoref{fig:embeddings:validationF1}, reveals a consistent gain of around $0.01$ on the validation set during fine-tuning.

\myparagraph{Conclusion} While \textsc{Clap} embeddings are powerful,  the synergy of \textsc{Clap}, \textsc{PalmTree}, and \textsc{Dexter} embeddings through \textsc{Combo} is the most effective design.

\subsubsection{\textsc{Lord}\label{sec:lordeva}}

\textsc{Lord} introduces a novel MLM task during fine-tuning and a confidence threshold during inference to enhance precision. Thus, we devise a version of \ProjectName named \textsc{Simple},  employing teacher forcing during fine-tuning with traditional left-to-right autoregression, but still with a confidence threshold. Additionally, we assess \textsc{Lord-T0}, a version of \textsc{Lord} where the confidence threshold is fixed at $0$. \textsc{Simple-T0} combines these two modifications. 
Lastly, as the \textsc{Combo} pre-training incorporates a captioning task, we also evaluate \textsc{Combo}'s direct predictions. 
We report summary results in~\autoref{tab:ablation:multi:testScores}. 
\textsc{Combo} as a decoder performs the worst across all metrics. This is expected; while \textsc{Combo} pre-training benefits from multiple objectives\mbox{~\cite{caruana1997multitask}}, conflicts and imbalances during optimization can hinder the performance of individual objectives\mbox{~\cite{sener2018multi}}. Fine-tuning is therefore essential for downstream tasks.

\begin{table}[t]
    \centering
    \newcommand{\spc}{\hspace*{7pt}}
    \begin{tabular}{l@{\spc}c@{\spc}c@{\spc}c@{\spc}c@{\spc}c@{\spc}c}
        \toprule
        Decoder & \makebox[0pt]{Prec.} & \makebox[0pt]{Recall} & \makebox[0pt]{$F_1$} & \makebox[0pt]{RougeL} & \makebox[0pt]{Bleu} & \makebox[16pt]{VarCLR\ }\\
        \midrule
         {\bf \textsc{Lord}} & {\bf 0.667} & 0.334 & {\bf 0.445} & 0.376 & 0.207 & 0.639 \\
        \textsc{Simple} & 0.471 & 0.379 & 0.420 & 0.400 & 0.235 & 0.651 \\
         \textsc{Lord-T0} & 0.436 & 0.377 & 0.404 & 0.397 & {\bf 0.261} & 0.642 \\   
        \textsc{Simple-T0} & 0.427 & {\bf 0.392} & 0.409 & {\bf 0.411} & 0.242 & \bf 0.653 \\
        \textsc{Combo} & 0.496 & 0.278 & 0.357  & 0.308 & 0.154 & 0.602\\
        \bottomrule
    \end{tabular}

\caption{\label{tab:ablation:multi:testScores}\textsc{Lord} ablation results. \textsc{Lord}: \ProjectName decoder, \textsc{Simple}: Standard left-to-right decoder, \textsc{-T0}: Confidence threshold is zero, \textsc{Combo}: \textsc{Combo} as a decoder.%
}
\end{table}

\textsc{Lord} achieves the highest precision at $0.667$  and $F_1$ score at $0.445$, demonstrating the effectiveness of both the novel MLM task and the confidence threshold despite a trade-off with the recall. In contrast, \textsc{Simple}, \textsc{Lord-T0}, and \textsc{Simple-T0} achieve better recall and RougeL scores, as RougeL is recall-related.
Moreover, \textsc{Simple-T0} achieves the highest VarCLR score at $0.653$ and \textsc{Lord-T0} the highest Bleu at $0.261$. Nevertheless, the contributions of \textsc{Lord} are positive as the $F_1$ score is the primary metric (see~\autoref{sec:metrics}).

\myparagraph{Conclusion}
 \textsc{Lord} allows an $F_1$ score gain of $0.036$ compared to the usual decoder architecture, thanks to a significant precision gain of $0.24$, which is essential in real-life scenarios. Moreover, \textsc{Combo} pre-training alone leads to weak results.

\begin{table*}[t]
\small
    \centering
    \begin{tabularx}{\linewidth}{cccXccccc}
        \toprule
         Case & Name &  Ground truth &  Model & Prediction & ${F_1}$ &  RougeL &  Bleu & VarCLR \\
        \midrule

&  &  & \ProjectName & \texttt{shell} & 0 & 0 & 0.841 & 0.521\\
&  &  & XFL  & \texttt{\_\_\_} & 0 & 0 & 0 & 0.564\\
\casebullet{1} &  \texttt{execute} & \texttt{execute}
& SymLM.  & \texttt{file\_main} & 0 & 0 & 0.639 & 0.413\\
&  &  & AsmDep.  & \texttt{string\_at} & 0 & 0 & 0.639 & 0.414\\
&  &  & HexT5  & \texttt{print\_sme\_za\_list} & 0 & 0 & 0.302 & 0.334\\
 \midrule

&  &  & \ProjectName & \texttt{pipe} & 0 & 0 & 0.114 & 0.336\\
& \texttt{eval} & \texttt{evaluate_} & XFL  & \texttt{\_\_\_} & 0 & 0 & 0 & 0.485\\
\casebullet{2} & \texttt{back} & \texttt{back_}
& SymLM  & \texttt{read} & 0 & 0 & 0.114 & 0.377\\
& \texttt{cmd}  & \texttt{cmd} & AsmDep. & \texttt{next\_window} & 0 & 0 & 0.388 & 0.408\\
&  &  & HexT5  & \texttt{new\_logical\_line} & 0 & 0 & 0.452 & 0.389\\
\midrule

&  &  & \ProjectName & \texttt{key\_file\_dump} & 0 & 0 & 0.452 & 0.427\\
& \texttt{fsa\_} &  & XFL  & \texttt{\_\_\_} & 0 & 0 & 0 & 0.417\\
\casebullet{3} & \texttt{get_} & \texttt{get_}
& SymLM  & \texttt{}\texttt{\_\_\_} & 0 & 0 & 0 & 0.417\\
& \texttt{config} & \texttt{config}  & AsmDep.  & \texttt{search\_done} & 0 & 0 & 0.639 & 0.381\\
&  &  & HexT5  & \texttt{madacc\_size} & 0 & 0 & 0.639 & 0.264\\ 
\midrule

&  &  & \ProjectName & \texttt{print\_date} & 0 & 0 & 0.639 & 0.594\\
&  &  & XFL  & \texttt{\_\_\_} & 0 & 0 & 0 & 0.536\\
\casebullet{4} & \texttt{dtws} & \texttt{time}
& SymLM  & \texttt{time} & 1 & 1 & 1 & 1\\
& \texttt{time} &  & AsmDep. & \texttt{cli\_color} & 0 & 0 & 0.639 & 0.321\\
&  &  & HexT5  & \texttt{reset\_items} & 0 & 0 & 0.639 & 0.361\\
\midrule

&  &  & \ProjectName & \texttt{text\_delete} & 0 & 0 & 0.639 & 0.696\\
& \texttt{remove} &  & XFL  & \texttt{\_\_\_} & 0 & 0 & 0 & 0.548\\
\casebullet{5} &  \texttt{From} & \texttt{remove_}
& SymLM  & function absent from the dataset & - & - & - & - \\
& \texttt{Edited} & \texttt{from} & AsmDep.  & \texttt{directory} & 0 & 0 & 0.309 & 0.437\\
&  &  & HexT5  & \texttt{\_collector\_env\_save\_preloads} & 0 & 0 & 0.302 & 0.361\\
\midrule

& \texttt{task} & \texttt{task_} & \ProjectName & \texttt{on\_button\_cancel\_clicked} & 0.444 & 0.444 & 0.368 & 0.756\\
& \texttt{panel\_} & \texttt{panel_} & XFL  & \texttt{\_\_\_} & 0 & 0 & 0 & 0.276\\
\casebullet{6} & \texttt{cancel_} & \texttt{cancel_}
& SymLM  & \texttt{}\texttt{\_\_\_} & 0 & 0 & 0 & 0.276\\
& \texttt{clicked_}
& \texttt{clicked_}
& AsmDep. & \texttt{\_\_\_}  & 0 & 0 & 0 & 0.276\\
&  \texttt{cbk} & \texttt{callback} & HexT5  & \texttt{gsl\_sf\_lnfact\_e} & 0 & 0 & 0.235 & 0.153\\
\bottomrule
\end{tabularx}
\caption{Case study of predictions in the cross-project setting. 
'\texttt{\_\_\_}': Empty prediction. 
 We report function-level $F_1$, RougeL, Bleu, and VarCLR scores. Bleu scores are surprisingly high for predictions that differ from the ground truth, because the smoothed version of BLEU-4\mbox{~\cite{lin-och04coling}} inflates n-gram precisions, which renders the score meaningless for very short sentences. In the same way, all of $F_1$, RougeL, and Bleu scores for HexT5 are affected by HexT5 employing a different vocabulary.}
\label{tab:project:casestudies}
\end{table*}

\section{Discussion\label{sec:discussion}}

The evaluation of \ProjectName shows significant improvement over state-of-the-art methods in both cross-binary and cross-project settings. Moreover, an ablation study confirms the relevance of our contributions to \ProjectName's success.

Despite these successes, the $F_1$ score of 0.460 in the cross-project setting is relatively low in absolute terms, warranting further investigation.
To investigate this issue further, we perform a qualitative case study on six instances where \ProjectName predictions diverged from the ground truth. This analysis aims to determine whether such predictions can be meaningful and contextually accurate despite not matching the ground truth. For completeness, we also include predictions from XFL, SymLM, AsmDepictor, and HexT5 in the same cases.
Predictions, ground truth, and function-level scores are shown in \autoref{tab:project:casestudies}.
We analyze the corresponding source code in each case to evaluate if the predictions are appropriate. For reference, we include source code for the functions in \autoref{appendix:casestudy}.

We first discuss the predictions of \ProjectName before briefly assessing those of other methods:
\pagebreak

\noindent
\casebullet{1} \ProjectName predicts \texttt{shell} instead of \texttt{execute}. The source code executes a shell script given in argument, therefore the prediction is close to the actual behavior of the function.

\noindent
\casebullet{2} \ProjectName predicts \texttt{pipe} instead of \texttt{evaluate\_back\_command}. Since the source code executes a command inside back quotes with a pipe, \ProjectName has captured a key functionality.

\noindent
\casebullet{3} \ProjectName predicts \texttt{key\_file\_dump} instead of \texttt{get\_config}. In the source code, the configuration returned is an instance of the GKeyFile structure. Hence, this prediction is more precise than the ground truth.

\noindent
\casebullet{4} \ProjectName predicts \texttt{print\_date} instead of \texttt{time}. Although the function does not print, it returns a structure akin to a datetime, making the prediction useful.

\noindent
\casebullet{5} \ProjectName predicts \texttt{text\_delete} instead of \texttt{remove\_from}. The original name of the function was \texttt{removeFromEdited}, but \texttt{edited} is not part of the vocabulary. This function comes from the hexedit project, a hexadecimal and ASCII text editor. The source code reveals that this function deletes text from an existing edition; therefore, the prediction is appropriate. Despite an $F_1$ score of 0, VarCLR validates the prediction with a score of $0.696$.

\noindent
\casebullet{6} \ProjectName predicts \texttt{on\_button\_cancel\_clicked} instead of \texttt{task\_panel\_cancel\_clicked\_callback}. Although different from the ground truth, this prediction is semantically close, as reflected by a VarCLR score of $0.756$. The name pattern \texttt{on...clicked} is common for callbacks on click events in graphical user interfaces, and the choice of \texttt{button} instead of \texttt{task\_panel} can be explained by the function's first argument being a \texttt{GtkButton} that has been clicked on.

In case six, BLens generates a clever rephrasing of the ground truth.
The ablation study data allows us to assess each component of BLens qualitatively.
Without \textsc{Combo} pre-training or with a single embedding, BLens predicts nothing.
Combining \textsc{Clap} with \textsc{PalmTree} results in \texttt{ui}, and with \textsc{Clap} and \textsc{Dexter} in \texttt{on\_delete\_clicked}.
A simpler decoder yields \texttt{on\_entry\_activate}.
This highlights the effectiveness of BLens's design.

Now, we briefly assess the other methods.
Firstly, XFL predictions are entirely empty, indicating that the model is conservative and avoids making incorrect predictions in challenging examples, resulting in more false negatives but fewer false positives. Secondly, among SymLM's five predictions, two are inappropriate based on the function names and source code (e.g., \casebullet{2}: \texttt{read} instead of \texttt{evaluate\_back\_command}), two are empty and one is correct. Note that in case five, the function is absent from SymLM's sub-dataset. Thirdly, AsmDepictor provides six predictions, of which five are inappropriate (e.g., \casebullet{4}: \texttt{string\_at} instead of \texttt{execute}), and one is empty. Lastly, all six predictions from HexT5 are inappropriate and meaningless (e.g., \casebullet{5}: \texttt{\_collector\_env\_save\_preloads} instead of \texttt{remove\_from}). A quick review of more predictions reveals that the pre-trained HexT5 model, which we used due to the unavailability of HexT5 training implementation, performs very poorly. This could also be due to large distribution shifts between datasets and low generalization from the model.

Overall, this case study suggests that, contrary to other methods, \ProjectName can predict meaningful names in the cross-project setting even if the $F_1$ score is low.
Nevertheless, there are some further points to discuss.

\myparagraph{Dataset overlap} Concerns about dataset overlap arise because we reused pre-trained \textsc{\textsc{Clap}} and \textsc{PalmTree} embeddings. However, these pre-trained models are based on broad datasets and trained with other learning objectives. Even if they have encountered some projects from our dataset, they are designed to generalize across different projects and compiler settings, rather than memorizing specific examples. Using pre-trained embeddings is common in machine learning as it leverages general-purpose models for new tasks, saving computational resources and enhancing model robustness for specific fine-tuning tasks. Nevertheless, BLens already surpasses the state of the art with only \textsc{Dexter} embeddings and merely 80 epochs\mbox{~(\autoref{tab:ablation:embeddings:testScores})}.

\myparagraph{Dataset diversity} Our dataset is taken from related research~\cite{acsac20-punstrip, oakland23-xfl}. This choice enables straightforward comparisons and grants a tokenizer specifically designed for this dataset. Since this dataset consists mostly of binaries from Debian packages written in C, the three thousand projects lack diversity. Nevertheless, \ProjectName could be applied to other datasets and programming languages.

\myparagraph{Evaluation metrics} 
Evaluating function names remains challenging due to the complexity of natural language.
VarCLR measures semantic similarity by embedding names to a high-dimensional space. This captures a degree of semantic similarity, as illustrated by cases five and six. Yet, VarCLR assigns high scores to empty predictions because the empty string lies at the center of the space. As seen in\mbox{~\autoref{tab:project:casestudies}}, VarCLR favors empty predictions over meaningful alternatives in the first two cases.
Additionally, incorrect predictions appear relevant (e.g., \texttt{directory} scores $0.437$ for \texttt{remove_from}), and VarCLR can be misaligned with human judgment, preferring \texttt{next_window} over \texttt{pipe} for \texttt{eval_back_cmd}.
In contrast, the $F_1$ score relies strictly on the set of words in the ground truth. Pre-processing of function names partially normalizes the ground truth, successfully addressing most abbreviations and smaller variations.

CodeWordNet\mbox{~\cite{jin22ccs,XiongASE23}} proposes considering word equivalences (e.g., \texttt{initialize} and \texttt{create}) during the $F_1$ computation. However, this requires manual validation and oversimplifies context-specific synonyms.
An alternative method is to use source code to better normalize function names, as proposed by Carvalho et al.~\cite{carvalho15systemsoftware}. For example, in the 3dchess project, \texttt{dir} would expand to \texttt{direction} rather than the more common \texttt{directory}. However, this requires access to the training data source code, which was not feasible in this study.

\section{Related Work\label{sec:relatedwork}}

\myparagraph{Function name prediction} Pioneering work on function name prediction relies on traditional machine learning techniques. For instance, Debin~\cite{he2018debin} and Punstrip~\cite{acsac20-punstrip} employ probabilistic graphical models,  while NRFE~\cite{han21issta} is a lightweight framework over various features. These approaches have low granularity and de facto identify functions frequent in the training data.
On the other hand, XFL~\cite{oakland23-xfl} predicts sets of words with multi-label classification. However, XFL's word ordering is agnostic to the binary code.
Recent works use the transformer~\cite{transformer} encoder-decoder architecture to translate binary code into function names, treating binary code and human-readable languages both as natural languages. For instance, AsmDepictor~\cite{kim23asiaccs} focuses on instructions, NERO~\cite{nero} uses augmented control flow graphs to represent the calling context, SymLM~\cite{jin22ccs} employs pre-trained function embeddings to capture execution behavior, and HexT5~\cite{XiongASE23} works with low-level pseudocode, though it relies on a challenging decompilation step to convert from binary code.

\myparagraph{Binary code information recovery} Variable name recovery and type inference require a more concrete representation of function behavior.
The pioneering work Debin~\cite{he2018debin} relied on memory cells and instructions to recover variable names. 
Some recent approaches rely on decompiled pseudocode.
DIRE~\cite{dire}, DIRECT~\cite{nitin2021direct}, and DIRTY~\cite{chen2022augmenting} use encoder-decoder architectures starting from pseudocode to recover variable names. The trend is toward designing multiple pre-training tasks~\cite{chen2022augmenting, XiongASE23}.
We already mentioned HexT5~\cite{XiongASE23}, which unifies multiple tasks, including summarization and function name prediction. These approaches rely on pseudocode, thus departing from binary code. CP-BCS~\cite{ye2023cpbcs} incorporates control flow graphs and assembly code but still primarily uses pseudocode.

\myparagraph{Source code information recovery} Efforts have also been made to recover source code information. Code2vec~\cite{code2vec} learns function embeddings from abstract syntax trees and can be fine-tuned for various tasks such as code retrieval and classification. Interestingly, it achieves a good $F_1$ score on function name prediction across projects. 
CodeT5~\cite{wang2021codet5} employs the transformer architecture to translate between function summaries and source code through various pre-training tasks. 
Source code presents different challenges than binary code because compiler optimizations introduce complex transformations and remove explicit types (e.g., unsigned int or array), making alignment harder.

\myparagraph{Embeddings for binary functions}
More approaches exist to represent binary functions. Xu et al.~\cite{gemini} use a Siamese architecture to train a graph neural network to represent control flow graphs, while Li et al.~\cite{li2019gmn} use a deep graph network and learn graph editing distances. They learn to discriminate pairs of binary functions compiled from different sources.
With the rise of transformer architectures, self-supervised tasks such as Masked Language Modeling have been adopted to learn deep semantics of assembly code~\cite{asm2vec, zhu2023ktrans, wang2022jtrans}.

\myparagraph{Image captioning} We have discussed several pre-training tasks, including Contrastive Image-Language Pre-training (CLIP)~\cite{clip} and contrastive captioning (CoCa)~\cite{coca}, which are essential to multimodal learning. Additional tasks have been developed, spanning unimodal tasks like image reconstruction~\cite{chen2020uniter} to multimodal tasks like object detection~\cite{zhang2021vinvl,li2020oscar}, word-region alignment~\cite{chen2020uniter}, and question answering~\cite{wang2022ofa}. 
Overall, the trend is toward combining multiple tasks in pre-training, a domain where OFA~\cite{wang2022ofa} excels.
However, defining equivalent tasks for binary code is difficult.
ViT~\cite{vit}, the original vision transformer, pioneered image patches and surpassed previous  convolutional approaches~\cite{kolesnikov2020big,xie2020self}.
Since then, other transformers have scaled to hundreds of millions of parameters, avoiding overfitting due to the generality of images and massive datasets~\cite{hu2022scaling, git}.

\myparagraph{Contrastive learning} 
Contrastive learning\mbox{~\cite{hadsell2006dimensionality}} minimizes the distance to anchors and maximizes it to others. We have already described techniques that employ text embedding as anchors, such as \textsc{Clap}\mbox{~\cite{WangISSTA24}} and \textsc{ContraBin}\mbox{~\cite{zhang2024contrabin}}. \mbox{\ProjectName} employs a learnable function name representation. Another line of work creates anchors with semantics-preserving transformations, but numerous transformations are required to obtain semantics-aware embeddings. Various transformations can be applied to source code\mbox{~\cite{jain2021contrastive, ding-etal-2022-towards, jiang2024bincola}}. Moreover, compiling the same source with different compilers, options, or for different architectures can produce anchors\mbox{~\cite{alphadiff,XiongASE23,sun2024graphmoco,jiang2024bincola}}.

\myparagraph{Embeddings fusion}
We use multimodal data fusion\mbox{~\cite{DeepMultimodalDataFusion}}  to incorporate different views of functions.
While fusing raw features\mbox{~\cite{fang2023featuresmultimodal, venugopalan2021multimodal}} might appear beneficial at first glance, fusing pre-trained embeddings takes advantage of the training and engineering efforts behind SotA embeddings.
To fuse heterogeneous data, 
cross-modal and intra-modal attention mechanisms\mbox{~\cite{hu2020bi, fang2023featuresmultimodal, gao2020multi, zhang2022can}}, as well as a transformer\mbox{~\cite{lei2021lessismore, chen2023xllm
}} can be applied.
An efficient attention process requires a sufficient number of tokens. Our innovation involves cutting function-level embeddings into multiple smaller patches, enabling the model to capture finer-grained relationships within the data. By incorporating positional encoding, we maintain the sequential integrity of the data.

\myparagraph{Transformer decoding}
There are several other decoding schemes apart from the classic left-to-right paradigm\mbox{~\cite{li2021otherautoregressiveorderings,qi20prophetnet, stern2019insertion, song2019mass, ghazvininejad2019mask, liao-etal-2020-probabilistically}}.
Our flexible autoregression scheme relies on Masked Language Modeling (MLM) to estimate the probability distribution of all unknown words given known words. Several parallel developments also utilize MLM to enhance speed\mbox{~\cite{ghazvininejad2019mask}} or to enable sentence generation in any order\mbox{~\cite{liao-etal-2020-probabilistically, song2019mass}}.

\section{Conclusion\label{sec:conclusion}}

We tackle the function name prediction problem by introducing \ProjectName, a novel approach inspired by multimodal learning.
It leverages two unique components: \textsc{Combo} and \textsc{Lord}.
\textsc{Combo} first ensembles state-of-the-art binary function embeddings into function patches. Then, patches are projected into tokens aligned with the word latent space via contrastive captioning loss, thereby capturing spatial relationships within the binary code. \textsc{Lord} decodes the function tokens into the function name through a new Masked Language Modeling task tailored specifically to function names.

\ProjectName sets a new state of the art in function name prediction, improving the main metric, $F_1$ score, by 12\%, with a precision increase to 0.917, which is critical for real-world applications. Additionally, \ProjectName achieves significant improvement in grammatical accuracy, with a 32.6\% boost in RougeL and a 79.1\% boost in Bleu scores. Against binaries coming from new projects, \ProjectName demonstrates remarkable robustness, increasing $F_1$ by 42.2\%, RougeL by 71.7\%, and Bleu by an impressive 188\%. Lastly, in a strict setting that reduces shared components BLens provides even more gains, with an improvement of $53.3$\% in terms of $F_1$ score.

\section*{Ethics Considerations}

The dataset used in this study comes from open-source Debian packages, and we did not perform any unauthorized gathering of binary code. On the one hand, we acknowledge that our work in reverse engineering could have negative implications for some stakeholders, including its potential misuse for copyright infringement.
On the other hand, advancing reverse engineering can significantly aid in identifying and mitigating malicious code early in an attack.
Overall, we believe that our work provides a net positive impact on software security.

\section*{Open Science}

We share the following artifacts on Zenodo\cite{blens-artifact}: BLens's implementation, pre-processed \textsc{Clap} embeddings, \textsc{PalmTree} embeddings, and \textsc{Dexter} embeddings for our dataset.

Due to GPL version 3~\cite{gplv3} licensing constraints, we cannot share the binary Debian packages dataset, as it requires conjointly distributing the source code. However, the Debian binaries are freely available separately. We believe that the shared artifacts will support the reproducibility and validation of our research findings.

\section*{Acknowledgments}
This work was funded in part by the Deutsche Forschungsgemeinschaft (DFG), reference \href{http://gepris.dfg.de/gepris/projekt/378803395}{378803395} (ConVeY)
and France Agence Nationale de la Recherche (ANR), program France 2030, reference ANR-22-PETCC-0001.

\bibliographystyle{abbrv}
\bibliography{bibliography}

\appendix

\section{Automatic Function Names\label{appendix:freebies}}

To accurately simulate a real-world scenario of using name prediction, e.g., within an advanced disassembler, we follow Patrick-Evans et al.~\cite{oakland23-xfl} by treating twenty functions as automatically inferred in each metric: 

\texttt{init}, \texttt{fini}, \texttt{csu\_init}, \texttt{csu\_fini}, \texttt{start}, \texttt{libc\_csu\_init}, \texttt{libc\_csu\_fini}, \texttt{libc\_start}, \texttt{deregister\_tm\_clones}, \texttt{register\_tm\_clones}, \texttt{rtld\_init}, \texttt{main}, \texttt{do\_global\_dtors\_aux}, \texttt{frame\_dummy}, \texttt{frame\_dummy\_init\_array\_entry}, \texttt{do\_global\_dtors\_aux\_fini\_array\_entry} \texttt{init\_array\_end}, \texttt{init\_array\_start}, \texttt{start\_main}, \texttt{libc\_start\_main}.
This list matches those functions automatically labeled by IDA Pro.

\section{Case Study\label{appendix:casestudy}} %

\begin{lstlisting}[style=CStyle, caption={Case \casebullet{1} source code.},captionpos=b]
int execute(char *program, char *action, 
            char *stardata) {
  char *path;	int result;
  path = malloc(strlen(SHELL)+strlen(SCRIPTS_DIR)+
             strlen(program)+strlen(action)+
             strlen(stardata)+4);
  if (!path) { return 1; }
  sprintf(path, "%s %s%s %s %s", SHELL,
       SCRIPTS_DIR, program, action, stardata);
  result = system(path); free(path);
  return result;	  
}
\end{lstlisting}

\begin{lstlisting}[style=CStyle, caption={Case \casebullet{2} source code.},captionpos=b]
void evalbackcmd(union node *n,
                 struct backcmd *result) {
  int pip[2]; struct job *jp;
  result->fd = -1; result->buf = NULL;
  result->nleft = 0; result->jp = NULL;
  if (n == NULL) { goto out; }
  if (pipe(pip)<0){sh_error("Pipe call failed");}
  jp = makejob(n, 1);
  if (forkshell(jp, n, FORK_NOJOB)==0) {
    FORCEINTON; close(pip[0]);
    if (pip[1] != 1) { dup2(pip[1], 1); 
                       close(pip[1]);  }
    ifsfree(); evaltreenr(n, EV_EXIT);
  }
  close(pip[1]);
  result->fd = pip[0]; result->jp = jp;
  out:
    TRACE(("evalbackcmd done: fd =%d buf=0x %x nleft=%d jp=0x%x\n", result->fd, result->buf, result->nleft, result->jp));
}
\end{lstlisting}

\begin{lstlisting}[style=CStyle, caption={Case \casebullet{3} source code.},captionpos=b]
GKeyFile *fsa_get_config(void) {
  GKeyFile *config_file = g_key_file_new();
  GError *err = NULL; 
  const char *filename = fs_get_config_file();
  int rv; int flags = G_KEY_FILE_NONE;
  rv = g_key_file_load_from_file(config_file, filename, flags, &err);
  if (!rv || err != NULL) {
    if (err->code != G_FILE_ERROR_NOENT && err->code != G_FILE_ERROR_EXIST && err->code != G_FILE_ERROR_ISDIR)
    {fsa_error(LOG_ERR, "error reading %s: %s(%d)", filename, err->message, err->code );}
    g_error_free(err);
    g_key_file_free(config_file);
    errno = ADM_ERR_GENERIC; return NULL;
  }
  errno = 0; return config_file;
}
\end{lstlisting}

\begin{lstlisting}[style=CStyle, caption={Case \casebullet{4} source code.},captionpos=b]
struct tws* dtwstime() { // from dtime.c
    long clock; (void) time( &clock );
    return dlocaltime( &clock ); 
}
struct tws { // from tws.h
    int tw_sec; int tw_min; int tw_hour;
    int tw_mday; int tw_mon; int tw_year; ... };
\end{lstlisting}

\begin{lstlisting}[style=CStyle, caption={Case \casebullet{5} source code.},captionpos=b]
void removeFromEdited(INT base, int size) {
  typePage *p, *q = NULL;
  for (p = edited; p; p ? (q = p, p = p->next) : (q = NULL, p = edited)) {
    if (base + size <= p->base) break;
    if (base <= p->base) {
      if (p->base + p->size <= base + size) {
        if (q) {q->next = p->next};
  	    else {edited = p->next};
  	    freePage(p); p = q;
      } else { p->size -= base+size - p->base;
               memmove(p->vals, p->vals + base
               + size - p->base, p->size);
               p->base = base + size; }
    } else if (p->base + p->size <= base + size) {
      if (base < p->base + p->size)
        {p->size -= p->base + p->size - base};
    } else { q = newPage(base + size, p->base
             + p->size - base - size);
             memcpy(q->vals, p->vals + base
             + size - p->base, q->size);
             q->next = p->next; p->next = q;
             p->size -= p->base + p->size - base;
             break; }
  }
  updatelastEditedLoc();
}
\end{lstlisting}

\begin{lstlisting}[style=CStyle, caption={Case \casebullet{6} source code.},captionpos=b]
int taskpanel_cancel_clicked_cbk(GtkButton *btn,
                                 gpointer data) {
  log_fct_enter();
  ui_taskpanel_update(current_task);
  log_fct_exit(); return FALSE; }
\end{lstlisting}

\end{document}